\documentclass{article}

\usepackage[final]{corl_2023} % Uncomment for the camera-ready ``final'' version.
\usepackage{graphicx}
\usepackage{wrapfig}
\usepackage{amsmath}
\usepackage{amsfonts}
\usepackage{cleveref}
\usepackage{enumitem}
\usepackage{comment}
\usepackage{subfig}
\usepackage{xspace}
\usepackage{xcolor}
\usepackage{tabularx}
\usepackage{bm}
\usepackage{algorithm}
\usepackage{algorithmic}
\usepackage{booktabs}
\usepackage{MnSymbol}
\usepackage{listings}

\usepackage{pythonhighlight}
\newcommand{\method}{CTG++\xspace}

\definecolor{tblue}{HTML}{1F77B4}
\definecolor{tred}{HTML}{FF6961}
\definecolor{tgreen}{HTML}{429E9D}
\definecolor{thighlight}{HTML}{000000}
\newcolumntype{P}{>{\raggedleft\arraybackslash}X}

\definecolor{cred}{HTML}{D62728}
\definecolor{cblue}{HTML}{1F77B4}
\definecolor{cgreen}{HTML}{79AB76}
\definecolor{cgrey}{rgb}{0.6,0.6,0.6}

\definecolor{highlight}{rgb}{0,0,0}

\newcommand{\bc}{\mathbf{c}}
\newcommand{\btau}[1]{\bm{\tau}^{#1}}

\newcommand{\bmu}{\boldsymbol{\mu}}
\newcommand{\bSigma}{\boldsymbol{\Sigma}}

\renewcommand{\bmu}{\bm{\mu}}

\renewcommand{\bSigma}{\bm{\Sigma}}

\newcommand\extrafootertext[1]{%
    \bgroup
    \renewcommand\thefootnote{\fnsymbol{footnote}}%
    \renewcommand\thempfootnote{\fnsymbol{mpfootnote}}%
    \footnotetext[0]{#1}%
    \egroup
}

\definecolor{codegreen}{rgb}{0,0.6,0}
\definecolor{codegray}{rgb}{0.5,0.5,0.5}
\definecolor{codepurple}{rgb}{0.58,0,0.82}
\definecolor{backcolour}{rgb}{0.95,0.95,0.92}

\newcommand{\new}[1]{\textcolor{black}{{#1}}}

\title{Language-Guided Traffic Simulation via \protect\\ Scene-Level Diffusion}

\author{
  Ziyuan Zhong$^1$, \ \ Davis Rempe$^{2,3}$,  \ \ Yuxiao Chen$^2$, \ \ Boris Ivanovic$^2$, \\ \textbf{Yulong Cao}$^2$, \ \ \textbf{Danfei Xu}$^{2,4}$, \ \ \textbf{Marco Pavone}$^{2,3}$, \ \ \textbf{Baishakhi Ray}$^1$ \\
  \\
  $^1$Columbia University, \quad $^2$NVIDIA Research, \quad $^3$Stanford University, \quad $^4$Georgia Tech
}

\begin{document}
\maketitle

%===============================================================================

\begin{abstract}
    Realistic and controllable traffic simulation is a core capability that is necessary to accelerate autonomous vehicle (AV) development. However, current approaches for controlling learning-based traffic models require significant domain expertise and are difficult for practitioners to use. To remedy this, we present \method, a scene-level conditional diffusion model that can be guided by language instructions. Developing this requires tackling two challenges:
    %Generating realistic traffic scenarios using a user-friendly interface, such as a textual one, can accelerate autonomous vehicle (AV) development. Although programmatic scenario description methods exist, they lack the intuitiveness of a natural language interface.  
    % To bridge this gap, we present \method, a language-guided, scene-level conditional diffusion model for scenario generation. Two key challenges emerge, 
    the need for a realistic and controllable traffic model backbone, and an effective method to interface with a traffic model using language.
    To address these challenges, we first propose a scene-level diffusion model equipped with a spatio-temporal transformer backbone, which generates realistic and controllable traffic. We then harness a large language model (LLM) to convert a user's query into a loss function, guiding the diffusion model towards query-compliant generation.
    %To address them, we propose a multi-agent diffusion model for realistic and expressive traffic modeling, and leverage a large language model (LLM) to translate a user query into a loss function to guide the diffusion model for query compliant generation. 
    % In order to capture complex interactions between agents, we further propose a multi-agent spatial-temporal transformer architecture for the diffusion model. 
    Through comprehensive evaluation, we demonstrate the effectiveness of our proposed method in generating realistic, query-compliant traffic simulations.
\end{abstract}

\keywords{Traffic Simulation, Multi-Agent Diffusion, Large Language Model} 

%===============================================================================

\section{Introduction}

% simulation testing -> importance of realism, controllability and user friendly interface -> what we want: realistic traffic simulation with text interface
Given the high costs and risks of large-scale real-world autonomous vehicle (AV) testing~\cite{adstats, nummilesneeded}, AV developers increasingly rely on simulations for developing robust systems~\cite{waymoreport}. For maximum efficacy, simulators must offer \textit{realistic} and \textit{controllable} traffic behaviors, complemented by a \textit{user-friendly interface}. The realism of traffic patterns ensures that development and testing conducted in simulation environments can be transferred to real-world scenarios. Controllability permits the generation of relevant traffic scenarios to scrutinize specific AV behaviors. For example, controlling a vehicle to collide with the AV to check how it reacts in dangerous situations. A user-friendly interface simplifies how desired behaviors can be specified. However, generating realistic~\cite{Suo_2021_CVPR,bits2022} and controllable~\cite{zhong2022guided} traffic poses considerable challenges, and the exploration of user-friendly interfaces in traffic generation has been limited. 
This work strives to develop an expressive scene-centric traffic model that can be controlled through a user-friendly text-based interface. %, in particular, a text-based interface.
%This work strives to develop a user-friendly interface for realistic traffic generation.
%A text-based 
Such an interface has the potential to connect simulation to previously unusable text-based data, such as governmental and insurance collision reports.
%Recognizing the high user-friendliness of text-based interfaces, we note their ability to unlock previously inaccessible data, such as governmental and insurance collision reports. 
It also facilitates new simulation capabilities, such as reconstructing real-world collision scenarios~\cite{nhtsa_crash_viewer}. 
%Given the potential for these exciting applications, in this work, we take the first step to developing a text-based interface for generating realistic traffic.

% Thus, our objective is to develop a text-based interface for generating realistic traffic.

% As a text-based interface is usually considered highly user friendly, enables harnessing previously-inaccessible data sources such as collision reports collected by governmental agencies and vehicle insurers, and supports new simulation capabilities such as reconstructing real-world collision scenarios \cite{nhtsa_crash_viewer}. 

%As text is the most natural way for communication and is the format for many good resources of challenging scenarios creation such as crash reports \cite{nhtsa_crash_viewer}, a model on realistic traffic simulation with a text-interface is thus desirable.

% 2 challenges; realistic and expressive traffic model; limitation of existing work
Building a traffic simulation model with a language interface presents two challenges. 
%realistic and expressive traffic modeling, and effective language-traffic model interfacing. 
First, the traffic model must generate realistic trajectories at both agent and scene levels, and provide controllability over its generated trajectories.
%and should possess the expressiveness to generate trajectories in response to a variety of language commands. 
Current simulators~\cite{sumo18, carla17, svl21}, whether replaying logs or using heuristic controllers for agent behavior, lack \textit{realism} and expressiveness. 
Data-driven approaches~\cite{Suo_2021_CVPR, bits2022} merely reflect training data distribution, lacking \emph{control} over generated traffic. 
Recently, CTG~\cite{zhong2022guided} applies a diffusion model, which has demonstrated promising results across various conditional generation tasks~\cite{dhariwal2021diffusion, li2022nlpdiffusion, janner2022diffuser, ajay2022conditional, rombach2022highresolution}, to traffic generation. 
CTG shows that diffusion is well-suited for controllable traffic simulation through \textit{guidance}, which allows test-time adaptability to user controls.
% This adoption is driven by the diffusion model's capabilities in distribution modeling and its test-time adaptability through a technique known as \emph{guidance}.
% Recently, CTG~\cite{zhong2022guided} applies a diffusion model, which has shown promising results in various conditional generation tasks~\cite{dhariwal2021diffusion, li2022nlpdiffusion, janner2022diffuser, ajay2022conditional, rombach2022highresolution}, into traffic generation due to its capability in distribution modeling and test-time adaptability through so-called \emph{guidance}. 
However, CTG models agents independently, leading to unrealistic interactions. For example, two vehicles modeled separately might collide if the leading vehicle slows without the following vehicle responding.
The second challenge is grounding language in a powerful traffic simulation backbone, since language conveys more abstract patterns (e.g., ``traffic jam" or ``following") while traffic models operate on low-level trajectories.
%Second, grounding language to a powerful traffic simulation backbone is challenging as the two have different modalities. 
% Considering the absence of a large-scale dataset with paired text and traffic data, it is infeasible to directly train a model to translate language into traffic patterns. 
To address similar issues, recent research on Large Language Models (LLMs) for robotic behaviors~\cite{lin2023text2motion, vemprala2023chatgpt} designs a suite of high-level functions (e.g., ``pick up" and ``use item") that an LLM can employ to control the robot in order to achieve a user-specified task (e.g., ``make an omelette"). Essentially, these high-level functions bridge textual instructions and robotic behaviors. Unfortunately, this approach cannot be directly used for realistic traffic simulation. It is infeasible for an LLM to only use a few high-level functions (e.g., ``go to location") to generate the entire low-level human-like trajectories.
% To address similar issues, recent research on Large Language Models (LLMs) for robotic behaviors~\cite{lin2023text2motion, vemprala2023chatgpt} typically designs a suite of high-level functions that an LLM can employ to control the robot in order to achieve a user-specified task. Essentially, these high-level functions act as an intermediary representation between textual instructions and robotic behaviors. Unfortunately, this approach cannot be directly implemented in a realistic traffic simulation. The difficulty lies in the ambiguity of designing a constrained set of functions for an LLM to utilize, such that these functions can generate realistic, high-dimensional trajectories that effectively mirror a user's intent.
%Unfortunately, such approach cannot be directly applied to realistic traffic simulation as it is unclear how to design only a few functions for a LLM to leverage such that those functions are enough to make the generated high-dimensional trajectories realistic and reflect a user's intent. 
%Reusing the APIs of a heuristic traffic simulator \cite{sumo18, carla17, svl21} does not fix its inherent limitation of lacking realism. Existing data-driven approach \cite{Suo_2021_CVPR, bits2022} does not provide an accessible interface supporting control via high-level functions. 

% solution to challenge1: multi-agent conditional diffusion; multi-agent spatial-temporal transformer. solution to challenge2: break down the language to traffic into two parts: lanugage to loss function (in code) and loss function to traffic (using diffusion).  

\begin{wrapfigure}{R}{0.6\textwidth}
\centering
\includegraphics[width=0.6\textwidth]{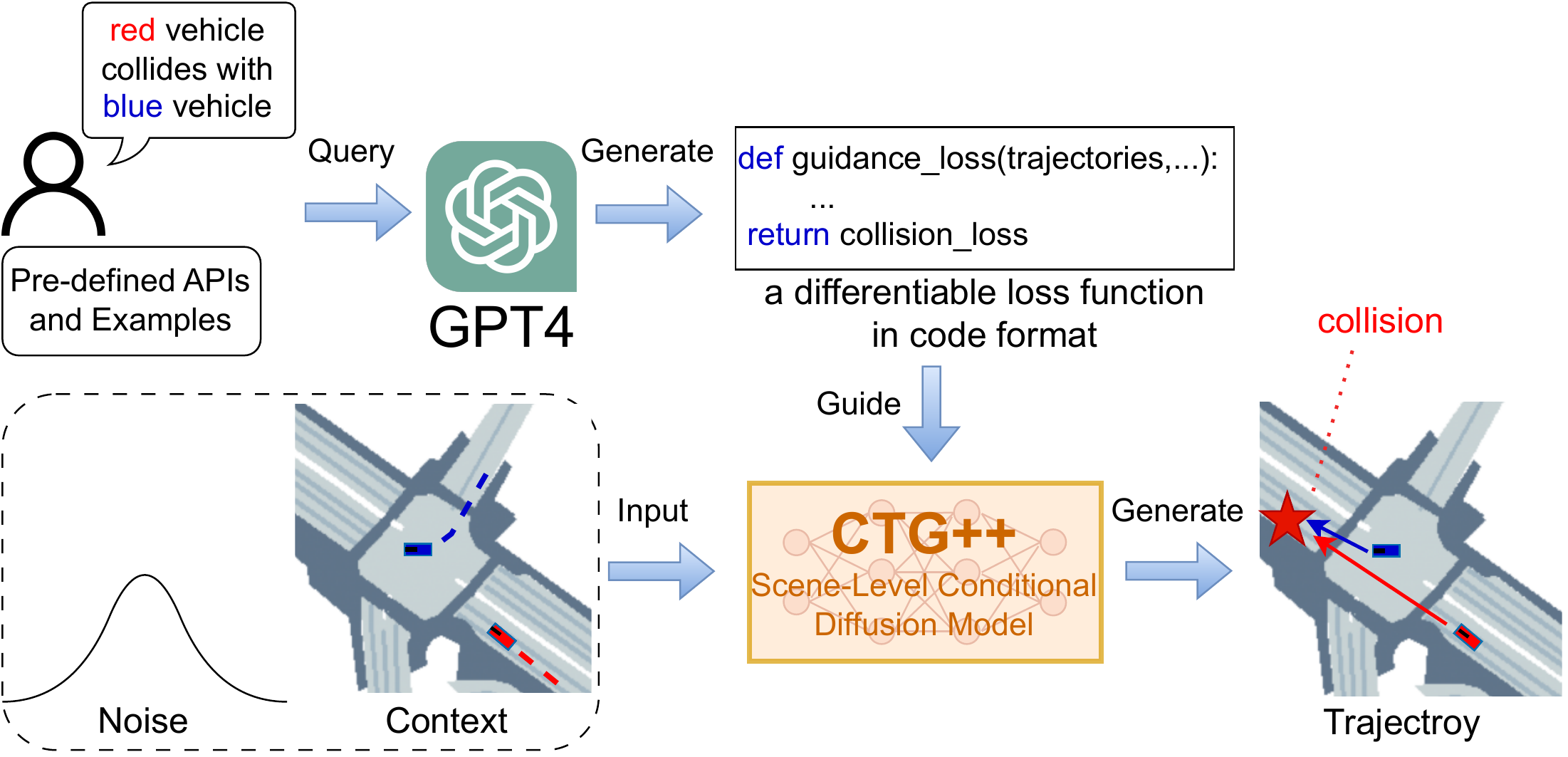}
\caption{\small{Overview of \method. A user query, predefined APIs, and examples are passed to GPT4, which generates a differentiable loss to guide \method for query-compliant trajectories.}}
\label{fig:overview}
\end{wrapfigure}

In this work, we propose a model called \method (see \Cref{fig:overview}) to overcome the aforementioned challenges. 
To achieve a realistic and adaptable traffic model, our approach harnesses the strengths of diffusion and significantly enhances its capability to cater to multi-agent scenarios. This is achieved through a newly proposed scene-level
% multi-agent 
conditional diffusion model, underpinned by a spatial-temporal transformer architecture. This architecture applies alternating temporal and spatial attention modules, effectively capturing the dynamics of multi-agent interactions over time. To create a natural language interface with the traffic model, we leverage the proven capacity of LLMs to generate code from natural language~\cite{bubeck2023sparks}. Instead of a direct translation from text to traffic, we introduce an intermediate representation: a differentiable loss function, which encodes a user's intention from the language command. The loss function guides the diffusion model to generate command-compliant trajectories. With this two-step translation, the LLM and diffusion model efficiently bridge the gap between user intent and traffic simulation. 
We evaluate \method on the nuScenes dataset \cite{nuscenes}, showing its ability to follow user-specified language commands and generate realistic trajectories. In summary, our contributions are:
\textbf{(1)} a scene-level conditional diffusion model, leveraging a spatial-temporal transformer backbone, designed for the generation of realistic and controllable traffic, \textbf{(2)} a language interface adept at generating trajectories that align with user-defined rules in language,
% \textbf{(1)} a language-guided multi-agent conditional diffusion model for generating realistic traffic that adheres to user-defined rules in natural language, 
and \textbf{(3)} extensive evaluation comparing \method to state-of-the-art baselines, highlighting its superiority in generating high-quality scenarios.

\section{Related Work}
\noindent\textbf{Traffic simulation.}
Traffic simulation methods can be divided into rule-based and learning-based. Rule-based strategies often feature an interface allowing users to specify the vehicles' routes, where the motion is governed by analytical models like intelligent driver model~\cite{microtrafficmodelbenchmark03}. Although they deliver user-friendly controllability, their behavioral expressiveness is limited, resulting in trajectories far from human driving patterns. To improve realism, learning-based approaches utilize deep generative models trained on trajectory datasets, aiming to emulate authentic driving behaviors~\cite{chai2019multipath,Chen_2022_CVPR,salzmann2020trajectron++, Suo_2021_CVPR, bits2022}. However, they trade off user-controllability for increased realism, as users cannot customize the properties of generated trajectories.
% Recent efforts have attempted to introduce controllability to learning-based approaches by directly optimizing generated trajectories, primarily for creating adversarial scenarios \cite{advdo, rempe2022strive}. However, they are restricted to adversarial settings and their direct optimization hurts realism.
% A recent work CTG~\cite{zhong2022guided} uses a diffusion model combined with Signal Temporal Logic (STL) guidance to generate realistic trajectories satisfying general objectives, such as stopping within specified regions. However, its STL interface necessitates manual adjustment for each new objective, posing scalability issues. 
In contrast, our scene-level diffusion model and LLM-based language interface allow the generation of realistic and language command-compliant traffic.

\noindent\textbf{Diffusion models for conditional generation in robotics.}
% Controllable diffusion models have been investigated using classifier-based~\cite{dhariwal2021diffusion}, classifier-free~\cite{ho2022classifier}, and reconstruction-based~\cite{ho2022video} guidance for image and video generation. 
As diffusion models have shown strong performance and test-time adaptability, they have been recently used for various conditional generation tasks in robotics and traffic. Existing works use trained classifiers \cite{janner2022diffuser, ajay2022conditional} or expert-designed loss functions \cite{zhong2022guided, rempe2023trace, MotionDiffuser} to guide the denoising process to achieve user-desired properties. For example, CTG~\cite{zhong2022guided}, the closest work to ours, uses a manually designed loss function based on Signal Temporal Logic (STL) to guide denoising.
% Diffuser \cite{janner2022diffuser} and DecisionDiffuser \cite{ajay2022conditional} use trained classifier(s) for guidance to plan robot behavior. Trace and Pace \cite{rempe2023trace} use domain specific loss functions for guidance to generate pedestrian animations. CTG \cite{zhong2022guided} uses traffic rules specified in STL loss for guidance to generate traffic simulations. A concurrent work MotionDiffuser \cite{MotionDiffuser} use expert designed loss functions for motion prediction. 
However, both training a classifier and manually crafting a loss function for each new property require domain expertise.
%are not scalable. 
In contrast, our approach allows a user to easily specify desired properties in natural language, which is then converted into a relevant loss function by an LLM. 
Moreover, most works, including CTG, model each agent independently \cite{Gu_2022_CVPR, janner2022diffuser, ajay2022conditional, zhong2022guided}, resulting in unrealistic agent interactions. In contrast, our scene-level diffusion model consider all agents in a scene jointly, resulting in realistic modelling of interactions.

% However, existing research primarily focuses on single-agent diffusion models \cite{Gu_2022_CVPR, janner2022diffuser, ajay2022conditional, zhong2022guided}, potentially resulting in sub-optimal generation when a scenario involves heavy multi-agent interactions.
% existing work all use a single-agent diffusion model \cite{Gu_2022_CVPR, janner2022diffuser, ajay2022conditional, zhong2022guided} and applies the same model into every agent which fails to capture the interaction of agents inherently in the diffusion model.

\noindent\textbf{Large language models for robotics.}
Recent breakthroughs in LLMs have motivated a series of works applying LLMs to robotic tasks. One approach is to train a multi-modal LLM that takes in embodied data in addition to text data \cite{driess2023palme, rt12022arxiv, ahn2022i}. Unfortunately, no such text-traffic data is available. Other works directly prompt a pre-trained LLM with a high-level function library along with a user query. This lets the LLM plan a robot's behaviors via the provided functions to achieve the goals in the query \cite{lin2023text2motion, vemprala2023chatgpt}. 
%This approach does not directly apply to traffic simulation due to the difficulty of using high-level functions to generate low-level human-like trajectories.
This approach does not directly apply to traffic simulation as existing data-driven approaches cannot be controlled via high-level functions. 
To tackle this challenge, we leverage a pre-trained LLM to translate a user query into a differentiable loss function in code format and use it to guide a scene-level conditional diffusion model for traffic generation.

\section{Methodology}
\label{sec:method}
After formulating the problem of controllable traffic generation (\Cref{sec:method-formulation}), we provide the details of our approach, \method. The training stage involves training a scene-level conditional diffusion model to capture diverse behaviors from real-world driving data (\Cref{sec:method-diffuser}), utilizing a scene-level spatial-temporal transformer architecture (\Cref{sec:method-diffuser-architecture}). During the inference stage, \method generates query-compliant behaviors via the guidance of a user query derived loss (\Cref{sec:method-guidance}). 
%Overall, the combination of the scene-level conditional diffusion model and language-based guidance enables the realistic and controllable generation of traffic trajectories in a user-friendly way.
% In this section, we provide the details of our approach, \method. \Cref{sec:method-formulation} formulates the problem of controllable traffic generation. We then outline the two stages of \method. The training stage involves training a scene-level conditional diffusion model to capture diverse behaviors from real-world driving data (\Cref{sec:method-diffuser}), utilizing our proposed scene-level spatial-temporal transformer architecture (\Cref{sec:method-diffuser-architecture}). During the inference stage, \method generates rule-compliant behaviors by sampling the model through a user query (in language) derived guidance loss (\Cref{sec:method-guidance}). Overall, the combination of the scene-level conditional diffusion model and language-based guidance enables the realistic and controllable generation of traffic trajectories in a user-friendly way.

\subsection{Problem Formulation}
\label{sec:method-formulation}
Similar to CTG~\cite{zhong2022guided}, we formulate the traffic simulation as an imitation learning problem. For the $M$ vehicles in a scene we want to simulate, let their state at a timestep $t$ be $s_{t}=[s_{t}^1~...~s_{t}^M]$ where $s_{t}^i = (x_{t}^i,y_{t}^i, v_{t}^i, \theta_{t}^i)$ (2D location, speed, and yaw) and the action (\textit{i.e.}, control) be $a_{t}=[a_{t}^1~...~a_{t}^M]$ where $a_{t}^i=(\dot{v}_{t},\dot{\theta}_{t})$ (acceleration and yaw rate). We denote $\bc=(I,s_{t-T_{hist}:t})$ to be decision-relevant context, which consists of local semantic maps for all the agents $I=\{I^1,...,I^M\}$, and their current and $T_{hist}$ previous states $s_{t-T_{hist}:t}=\{s_{t-T_{hist}}, \dots, s_{t}\}$.
To obtain state $s_{t+1}$ at time $t+1$, we assume a transition function (e.g., a unicycle dynamics model) $f$ that computes $s_{t+1}=f(s_{t}, a_{t})$ given the previous state $s_{t}$ and control $a_{t}$. 
Our goal is to generate realistic and query-satisfying traffic behavior for the agents given (1) the decision context $\bc$ and (2) a function $r : \mathbb{R}^{4T} \times \mathbb{R}^{2T} \rightarrow \mathbb{R}$ derived from a user query to measure rule satisfaction of the state and action trajectories. A model should generate future trajectories for the agents $s_{t+1:t+T}$ over the next $T$ timesteps. Ideally, these trajectories maximize satisfaction $r(a_{t:t+T-1},s_{t+1:t+T})$ to avoid violating the given rule.

% However, in many cases there is an inherent tradeoff between rule satisfaction and trajectory realism: \textit{e.g.}, if a user seeks a simulated vehicle with a speed much slower/faster than the speed limit, this is naturally ``unrealistic" in the context of city streets. 
% Therefore we must strike a balance between meeting user-specified constraints while still maintaining realistic behavior.
% As described next, our method does this by first training a rule-agnostic traffic generation model on real-world data to capture \textit{realism}, which is then guided for rule-specific compliance only during inference. 

\subsection{Scene-Level Conditional Diffusion for Traffic Modeling}
\label{sec:method-diffuser}
Diffusion models \cite{pmlr-v37-sohl-dickstein15,NEURIPS2020_4c5bcfec,janner2022diffuser, ajay2022conditional} generate new samples through an iterative denoising process by learning to reverse a diffusion process. As a traffic scene involves multiple traffic participants, a single-agent diffusion model \cite{janner2022diffuser, ajay2022conditional, zhong2022guided} may generate sub-optimal samples when a scene involves significant interactions among multiple agents. To tackle this issue, we propose a scene-level diffusion model that jointly models all traffic participants in a scene.
Unlike CTG, which models each agent's future trajectory independently, our model operates on the past and future trajectories of all the agents in a scene jointly (see \Cref{fig:state_action}) and thus captures the interactions among agents both spatially and temporally.
%As shown in \Cref{fig:state_action}, our diffusion model operates on future trajectories of states and actions~\cite{janner2022diffuser, zhong2022guided} and is conditional on the context (including history trajectories and map) at each step of denoising. 
Starting from Gaussian noise, the diffusion model is applied iteratively to predict a clean, denoised trajectory of states and actions for all agents in a scene.

\noindent\textbf{Trajectory Representation.} We denote the future trajectory that the model operates on as:
\begin{equation*}
    \btau{} := \begin{bmatrix}
            \btau{}_s  \\
            \btau{}_a  \\
        \end{bmatrix}, \quad 
        \btau{}_a := \begin{bmatrix}
            \btau{1}_a  \\
            \vdots \\
            \btau{M}_a  \\
        \end{bmatrix}, \quad 
        \btau{}_s := \begin{bmatrix}
            \btau{1}_s  \\
            \vdots \\
            \btau{M}_s  \\
        \end{bmatrix}, \quad 
        \btau{i}_a := [a_0^i~...~a_{T-1}^i], \quad 
        \btau{i}_s := [s_1^i~...~s_{T}^i].
\end{equation*}
Additionally, we represent the historical trajectory in the context $\bc$ as $\btau{}_{hist}$. In accordance with \cite{zhong2022guided}, our model predicts solely the action trajectory $\btau{}_a$, employing the known dynamics $f$ to deduce the states $\btau{}_s$ via a rollout from the initial state $s_0$ (which forms part of the context $\bc$). To maintain the physical feasibility of the state trajectory throughout the denoising process, we consistently denote $\btau{}_s$ as a state trajectory resulting from actions: $\btau{}_s = f(s_0, \btau{}_a)$.
% We additionally denote the history trajectory in the context as $\btau{}_{hist}$. Similar to \cite{zhong2022guided}, our model predicts only the actions $\btau{}_a$, and utilize the known dynamics $f$ to derive the states $\btau{}_s$ via a rollout starting from the initial state $s_0$ (part of the context $\bc$). To ensure the physical feasibility of the state trajectory throughout the denoising process, in the following, we always denote $\btau{}_s$ to be a state trajectory resulting from actions: $\btau{}_s = f(s_0, \btau{}_a)$.

\noindent\textbf{Formulation.}  Consider $\btau{k}_a$ as the action trajectory at the $k$-th diffusion step, with $k=0$ marking the original clean trajectory.
The forward diffusion process that starts from $\btau{0}_a$ is defined as:
\begin{equation}
\setlength{\abovedisplayskip}{1pt}
\setlength{\belowdisplayskip}{1pt}
\label{eq: diffusion_q} q(\btau{1:K}_a|\btau{0}_a) :=  \prod_{k=1}^{K} q(\btau{k}_a|\btau{k-1}_a) := \prod_{k=1}^{K} \mathcal{N}(\btau{k}_a; \sqrt{1-\beta_{k}}\btau{k-1}_a,\beta_{k}\mathbf{I}).
\end{equation}
Here, $\beta_{i}$ for all $i=1,...,K$ are a pre-determined variance schedule, controlling the magnitude of noise added at each diffusion step. As the noise incrementally accumulates, the signal transforms into an approximately isotropic Gaussian distribution $\mathcal{N}(\mathbf{0}, \mathbf{I})$.
For trajectory generation, our goal is to reverse this diffusion process through a learned conditional denoising model (\Cref{fig:state_action}) that is iteratively applied starting from sampled noise.
The reverse diffusion process,
%conditioned on the context information $\bc$, 
is as follows:
\begin{equation}
\setlength{\abovedisplayskip}{1pt}
\setlength{\belowdisplayskip}{1pt}
\label{eq: denoise_p} 
p_{\theta}(\btau{0:K}_a | \bc) :=  p(\btau{K}_a)\prod_{k=1}^{K} p_{\theta}(\btau{k-1}_a|\btau{k}, \bc) := p(\btau{K}_a)\prod_{k=1}^{K} \mathcal{N}(\btau{k-1}_a; \boldsymbol{\mu}_{\theta}(\btau{k},k,\bc),\boldsymbol{\Sigma}_{\theta}(\btau{k},k,\bc)),\\
\end{equation}
where $p(\btau{K}_a) = \mathcal{N}(\mathbf{0}, \mathbf{I})$ and $\theta$ represents the parameters of the diffusion model.
At each step, the model takes in actions $\btau{k}_a$ and the resulting states $\btau{k}_s = f(s_0, \btau{k}_a)$ as input.
As per ~\cite{NEURIPS2020_4c5bcfec, Gu_2022_CVPR}, the variance term of the transition is a fixed schedule such that $\boldsymbol{\Sigma}_{\theta}(\btau{k},k,\bc) = \boldsymbol{\Sigma}^k = \sigma_{k}^2\mathbf{I}$. The training in our approach mirrors that in \cite{zhong2022guided}, but with a key difference - trajectories are sampled at the scene level instead of the agent level, as our model simultaneously predicts outcomes for all agents in a scene. Detailed information is available in \Cref{sec:algo-training}.

\subsection{Model Architecture: Scene-Level Spatial-Temporal Transformer}
\label{sec:method-diffuser-architecture}
\begin{figure}[t!]
\centering
\includegraphics[width=\textwidth]{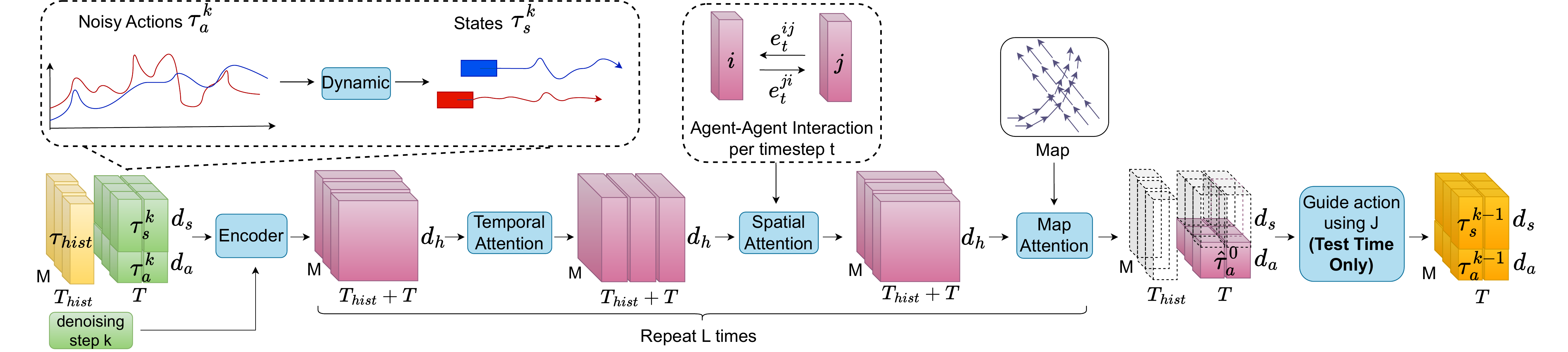}
\caption{\small{A denoising step using our scene-level spatial-temporal transformer. 
%$d_s$, $d_a$, and $d_h$ represent the dimensions of action, state, and latent for each vehicle per timestep.
}}
\label{fig:state_action}
\end{figure}

Unlike previous works which use a U-Net~\cite{janner2022diffuser, zhong2022guided} or single-agent transformer \cite{Gu_2022_CVPR} to model the denoising process for a single agent, we design an architecture that models multiple agents jointly. Inspired by recent works on transformer-based motion prediction \cite{yuan2021agent, ngiam2022scene, girgis2022latent, Zhou_2022_CVPR, nayakanti2022wayformer}, we propose a spatial-temporal transformer architecture to model multiple agents jointly. Unlike most previous work \cite{yuan2021agent, ngiam2022scene}, which employs scene-centric coordinates to capture interactions, we adopt agent-centric coordinates.
As a traffic scene configuration is combinatorial, it is easy for a scene to end up in an out-of-distribution configuration if it is modeled in scene-centric coordinates, since errors compound over time.
% ground truth traffic patterns differ from prediction as compounding errors accumulate over a rollout, and scene-centric coordinates can quickly lead to an out-of-distribution context. 
In contrast, agent-centric coordinates are invariant to translation and rotation of the scene,
%equivariant 
and therefore more robust during closed-loop simulation.
% rollout. 
However, agent-centric coordinates discard relative information among agents, which is important for interactions. To avoid this, we introduce a spatial attention module that enables the exchange of relative information between agents. Inspired by previous work~\cite{nayakanti2022wayformer}, to avoid the combinatoric explosion of attention pairs, we alternate between temporal attention, spatial attention, and map attention module to fully capture the interactions among agents and the map. We next introduce the details of our proposed 
architecture by showing the data flow of a denoising step (\Cref{fig:state_action}; see \Cref{fig:state_action_detailed} for more details).

\noindent\textbf{Input and Temporal Attention.} We first concatenate the ground-truth agent history trajectories with the predicted future trajectories along the temporal dimension and apply a row-wise feed-forward network (rFFN) to project each element (each agent per timestep) from the attribute dimension $d_s+d_a$ to the hidden dimension $d_h$. The denoising step $k$ is injected into the encoded trajectory using a sinusoidal positional encoding function \cite{transformer2017}. We next capture the temporal information in the encoded trajectory by feeding it into the temporal attention block, a standard transformer encoder~\cite{transformer2017} that captures the temporal-wise relation of each agent.
% interaction of each agent. 

\noindent\textbf{Spatial Attention.}
The encoded trajectory is then fed into a spatial attention block which is a customized transformer decoder block with key and value designed to capture the \textit{relative} geometric relationships among agents. Similar to \cite{Zhou_2022_CVPR}, we extend a regular attention layer to be aware of the relative information $\mathbf{e}_t^{ij}$ between two agents $i$ and $j$ at timestep $t$:
% \begin{equation}
% \mathbf{e}_t^{ij}=\phi_{\mathrm{rel}}\left(\left[\mathbf{R}_t^{i^\top}\left((\Delta x_t^{i j}, \Delta y_t^{i j})\right), \cos (\Delta \theta_t^{i j}), \sin (\Delta \theta_t^{i j}), v_t^j \cos (\Delta \theta_t^{i j})-v_t^i, v_t^j \sin (\Delta \theta_t^{i j})\right]\right)
% \end{equation} 
\begin{equation}
\mathbf{e}_t^{ij}=\phi_{\mathrm{r}}\left(\left[\mathbf{R}_0^{i^\top}\left( \Delta x_{0,t}^{i j}, \Delta y_{0,t}^{i j}\right), \cos (\Delta\theta_{0,t}^{i j}), \sin (\Delta\theta_{0,t}^{i j}), v_t^j \cos (\Delta\theta_{0,t}^{i j}) - v_0^i, v_t^j \sin (\Delta\theta_{0,t}^{i j}), d_t^{i,j}\right]\right)
\label{eq:edge}
\end{equation}
where $\phi_{\mathrm{r}}$ is a feed-forward network, $\Delta x_{0,t}^{i j}:=x_t^{j}-x_0^{i},\Delta y_{0,t}^{i j}:=y_t^{j}-y_0^{i},$ and $\Delta \theta_{0,t}^{i j}:=\theta_t^{j}-\theta_0^{i}$ represent the position and orientation differences from $j$ at timestep $t$ to $i$ at time step $0$ (the current timestep), $d_t^{ij}$ is the relative distance between $i$ and $j$ at timestep $t$, and $\mathbf{R}_0^{i^\top}$ is the rotation matrix associated with agent $i$ at timestep $0$.
For future timesteps at the training stage and history timesteps, we use the ground-truth relative information. For future timesteps at the inference stage, since we do not have the ground-truth information, we use a constant velocity model (which assumes the agents to keep their current velocity for the future timesteps as in \cite{scept}) to estimate the state of all the agents in the future and thus their relative information. 
% while for future timesteps we use a constant velocity model (which assumes the agents to keep their current velocity for the future timesteps as in \cite{scept}) to estimate the state of all the agents in the future and thus their relative information. 
The pair-wise relative information is then incorporated into the transformation of the encoded trajectory via keys and values of the decoder layer:
\begin{equation}
\mathbf{q}_t^i =\mathbf{W}^{Q^{\text {global }}} \mathbf{h}_t^i, \quad
\mathbf{k}_t^{i j} =\mathbf{W}^{K^{\text {global }}}\left[\mathbf{h}_t^j, \mathbf{e}_t^{i j}\right], \quad
\mathbf{v}_t^{i j} =\mathbf{W}^{V^{\text {global }}}\left[\mathbf{h}_t^j, \mathbf{e}_t^{i j}\right]
\end{equation}
where $h_t^i$ and $h_t^j$ are the slices of the encoded trajectories corresponding to the
%local embeddings
agent $i$ and $j$ at timestep $t$, and $\mathbf{W}^{Q^{\text{global}}}$, $\mathbf{W}^{K^{\text{global}}}$,
$\mathbf{W}^{V^{\text{global}}}$ are learnable matrices.

\noindent\textbf{Map Attention and Output.}
The map attention block is a multi-head attention layer with keys and values from the encoded agent-centric vectorized map (as in \cite{girgis2022latent}, we encode the map via an attention layer which transforms waypoints associated with each lane into a lane vector)
% (we follow \cite{girgis2022latent} for map encoding) 
and captures the interaction between agents and map. The map attention is applied to each agent independently as the map is agent-centric. The output encoded trajectory is projected back to the input dimension $d_s+d_a$ and results in the predicted clean action trajectory $\hat{\boldsymbol{\tau}}_a^0$. At test time, we additionally apply iterative guidance with a differentiable loss function $\mathcal{J}$ (see \Cref{sec:method-guidance} and \Cref{sec:algo-sampling}) on the predicted action trajectory. Finally, we apply dynamics to get the predicted state trajectory.

% The input context $\mathbf{c}$ containing agent-centric map information and past trajectories is represented in a rasterized format. This context is processed by a ResNet~\cite{he2016deep} encoder $\mathcal{F_{\theta}}$ before being passed to the diffusion model. 
% Similar to Diffuser \cite{janner2022diffuser}, we use a diffusion model architecture like U-Net containing several blocks of temporal 1D convolutions over the input trajectory. 
% We incorporate conditioning information by first concatenating with the diffusion step input $k$ and then adding this conditioning feature to the convolutional features at each block of the U-Net.
% The diffusion process uses a cosine variance schedule~\cite{janner2022diffuser,nichol2021improved} and $K=100$ diffusion steps for all experiments.

\subsection{Guided Generation with Language}
\label{sec:method-guidance}
A language interface for the powerful diffusion model would enable the user to easily control trajectories with minimum domain knowledge. However, the absence of paired text-to-traffic data renders direct training of such a model infeasible. Hence, we explore using an intermediary representation to bridge the two. 
Recent advancements in LLMs facilitate high-quality conversion from natural language commands into code. Meanwhile, the diffusion model exhibits control over its generation through guidance from a loss function. Thus, we suggest utilizing a loss function implemented in code to bridge the two.
% Given the recent advancements in LLMs, which enable high-quality conversion from natural language commands to code, and the diffusion model's controllable generation capability via guidance from a loss function, we propose using a coded loss function as the bridging entity. 
Since the guidance loss function must operate on trajectories, we provide helper functions for coordinate transformations and a handful of text-loss function paired examples alongside the user's query to the LLM. We then utilize the returned loss function to guide the diffusion model, as discussed in \Cref{sec:method-diffuser}, for generating query-compliant traffic simulation.
% To enforce desired rules on realistic samples from the trained diffusion model in a scalable and user-friendly way, we use an iterative guidance algorithm with rules specified in natural language. The language interface enables the user to easily control the trajectories with minimum domain knowledge and potentially reconstruct challenging scenarios from crash reports. As diffusion model requires a differentiable classifier to guide trajectory generation, a reward/loss function (with respect to predicted trajectories) that can capture the intention in a user's query needs to be constructed. 
% As such a loss function involves the manipulation of trajectories involving real-world physics, a text-based LLM may not inherently possess such capability. To address this challenge, we provide helper functions involving coordinate transformation and a couple of query-loss function paired examples in addition to a user's query to a LLM, and use the returned loss function to guide the diffusion model (discussed in \Cref{sec:method-diffuser}) to generate query-compliant traffic.

\noindent\textbf{Guidance Formulation.}
Building upon prior work~\cite{janner2022diffuser, zhong2022guided}, we apply \textit{guidance} to sampled trajectories from our
% conditional 
diffusion model at each denoising step to satisfy a predefined objective. Guidance uses the gradient of the loss $\mathcal{J}$ to perturb the model's predicted mean such that each denoising step (in \Cref{eq: denoise_p}) becomes:
$p_\theta(\btau{k-1}_a \mid \btau{k}, \bc) \approx \mathcal{N}(\btau{k-1}_a; \bmu + \bSigma^k \nabla_{\boldsymbol{\mu}} \mathcal{J}(\boldsymbol{\mu}) , \bSigma^k)$ (see \Cref{sec:algo-sampling}).

\noindent\textbf{Language Guidance.}
\begin{figure}[t!]
\centering
\includegraphics[width=\textwidth]{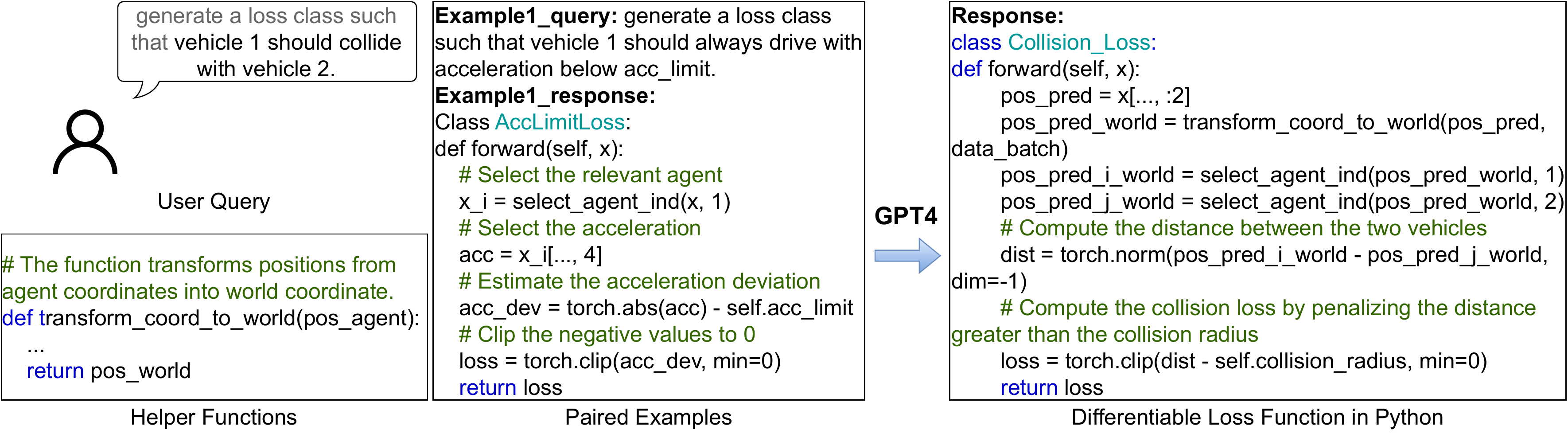}
\caption{\small{Example of prompting and querying LLM for a loss function promoting two vehicles to collide.}}
\label{fig:gpt_example}
\end{figure}
Rather than training a classifier or reward function \cite{janner2022diffuser, li2022nlpdiffusion} or defining an analytical reward function \cite{zhong2022guided} for $\mathcal{J}$, we use GPT4~\cite{openai2023gpt4} to translate the intention in a user language query into the corresponding guidance function. In particular, we pass a few helper functions used to manipulate trajectory coordinates, a couple of (query $\rightarrow$ loss function code) paired examples, and the user query into GPT4, and extract an implemented loss function from the returned message. 
% As illustrated in \Cref{fig:overview}, for each user queried rule we aim to enforce through guidance,  GPT4 produces a corresponding guidance function $\mathcal{J}$ in Python which guides the diffusion model for query-compliant trajectory generation. 
\Cref{fig:gpt_example} shows an example of querying for a loss function that causes a vehicle to collide with another one. The input paired example shows GPT4 how to apply a sequence of differentiable operations to generate a loss with respect to the predicted trajectory (``x"). 
%The loss function (``acceleration limit") becomes larger if the predicted trajectory violates a user's query (i.e., the trajectory has acceleration larger than the acceleration limit).
%that penalizes any violation of the intention from the user's query (``acceleration limit").
We also provide a list of helper functions for trajectory manipulation (such as coordinate transformations) for GPT4 to leverage as we have found these can help to avoid minor mistakes on common operations (see \Cref{sec:gpt-prompt} for additional helper functions and pair examples). GPT4 returns a loss function that penalizes trajectories for not having a collision between the specified vehicles, which will guide the diffusion model to generate trajectories having such a collision. 
% Our approach is scalable as the user's query in language is much simpler than carefully designed STL formulas \cite{zhong2022guided}. 
% When multiple rules are applied, we sum their corresponding losses to form $\mathcal{J}$. 
The guided sampling is performed jointly for all the agents in a scene.

\noindent\textbf{Traffic Simulation.}
We perform closed-loop traffic simulation of each scene for $10$ seconds. In particular, we 
apply our model for all the agents in a standard control loop: at each step, the model generates a guided trajectory and takes the first few actions before re-planning at $2$ Hz.
\section{Experiments}
\label{sec:experiments}
Following the setup (\Cref{sec:setup}), we conduct experiments to affirm: \method can effectively produce realistic and query-compliant traffic behaviors (\Cref{sec:experiment-gpt_case_study}), and compared with strong baselines, \method yields superior trade-offs among stability, rule satisfaction, and realism (\Cref{sec:experiment-traffic_model_evaluation}). 

% We conduct experiments to affirm: (1) \method can effectively produce realistic and query-compliant traffic behaviors, and (2) compared with strong baselines, the scene-level conditional diffusion model employed by \method yields superior trade-offs among stability, rule satisfaction, and realism. 
% Following the setup in \Cref{sec:setup}, we undertake two qualitative case studies utilizing \method for text-to-traffic simulation, demonstrating the effectiveness of the language interface (\Cref{sec:experiment-gpt_case_study}). We then assess the traffic model component of \method against baselines under rules generated by GPT and STL rules from \cite{zhong2022guided}, and perform an ablation study on design choices (\Cref{sec:experiment-traffic_model_evaluation}).

\subsection{Experimental Setup}
\label{sec:setup}
\noindent\textbf{Dataset.} 
nuScenes \cite{nuscenes} is an extensive real-world driving dataset encompassing 5.5 hours of precise trajectories from two cities, featuring diverse scenarios and heavy traffic. We train all models using the training split and evaluate them on 100 scenes randomly selected from the validation split. Our focus is exclusively on simulating moving vehicles, as they are the most control-relevant entities.
%In this study, we concentrate solely on simulating vehicles and apply the models only to vehicles with non-zero speed in the dataset as they are the interesting ones to control.

\noindent\textbf{Metrics.}
% Our evaluation contains two parts: language interface and traffic model. For the first part, we conduct two qualitative case study of using the language interface for text to traffic simulation.
% For the second part, 
Following \cite{zhong2022guided, bits2022}, we evaluate stability (\textit{i.e.}, avoiding collisions and off-road driving), controllability, and realism of generated trajectories. We evaluate \textit{stability} by reporting the failure rate (\textbf{fail}), measured as the percentage of agents encountering a collision or road departure in a scene. We evaluate \textit{controllability} using rule-specific violation metrics (\textbf{rule}) (see \Cref{sec:experiment-metrics_rule}). To assess \textit{realism}, we compare data statistics between generated trajectories and ground truth trajectories from the dataset by calculating the Wasserstein distance between their normalized histograms of driving profiles. We measure \textit{realism} using realism deviation (\textbf{real}), which is the average of the distribution distance for the longitudinal acceleration magnitude, lateral acceleration magnitude, and jerk. We introduce a new scene-level realism metric (\textbf{rel real}) which is the average of the distribution distance for relative (averaged over every pair of vehicles in a scene) longitudinal acceleration magnitude, relative lateral acceleration magnitude, and relative jerk. 

\noindent\textbf{Traffic Model Baselines.} 
The closest related work on rule-compliant traffic generation is \textbf{CTG} \cite{zhong2022guided}, a traffic model based on conditional diffusion. We also consider \textbf{BITS} \cite{bits2022}, a bi-level imitation learning model, and adapt its sampling ranking function to use our loss function. Its variant, \textbf{BITS+opt}, applies optimization to the output action trajectory for controllability. To ensure a fair comparison, this optimization employs the same loss function as the one used for guidance in \method.

\noindent\textbf{Large Language Model.} We use OpenAI's \textit{GPT-4} \cite{openai2023gpt4} (accessed through the OpenAI API) for evaluating the language interface. We do not train the LLM and use only few-shot prompting.

\subsection{Case Study of Language Interface}
\label{sec:experiment-gpt_case_study}

\begin{figure}[t]
\centering
\subfloat[][\small{\method keep distance} \label{fig:gpt_example-ctg++-keepdistance}]{\includegraphics[width=0.24\textwidth]{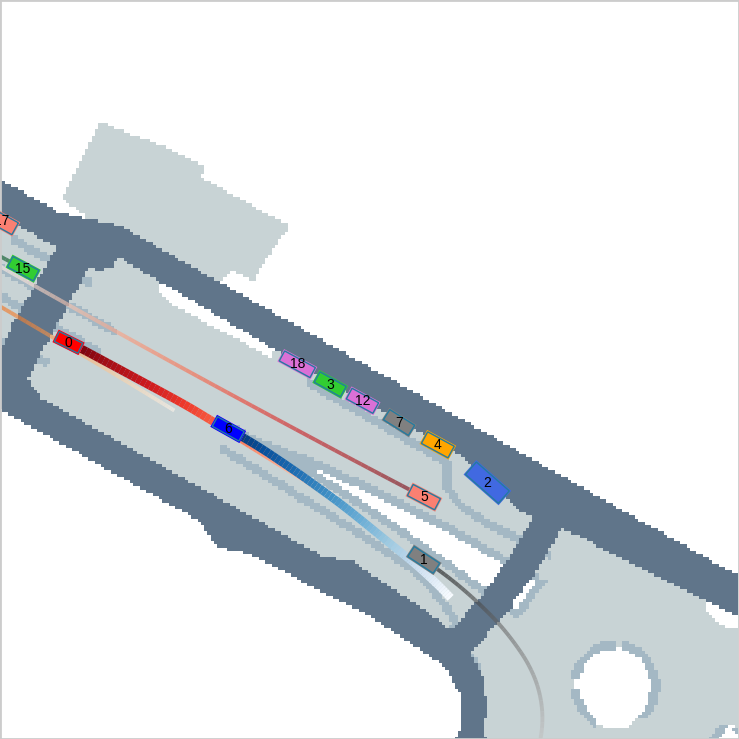}}
\subfloat[][\small{CTG keep distance} \label{fig:gpt_example-ctg-keepdistance}]{\includegraphics[width=0.24\textwidth]{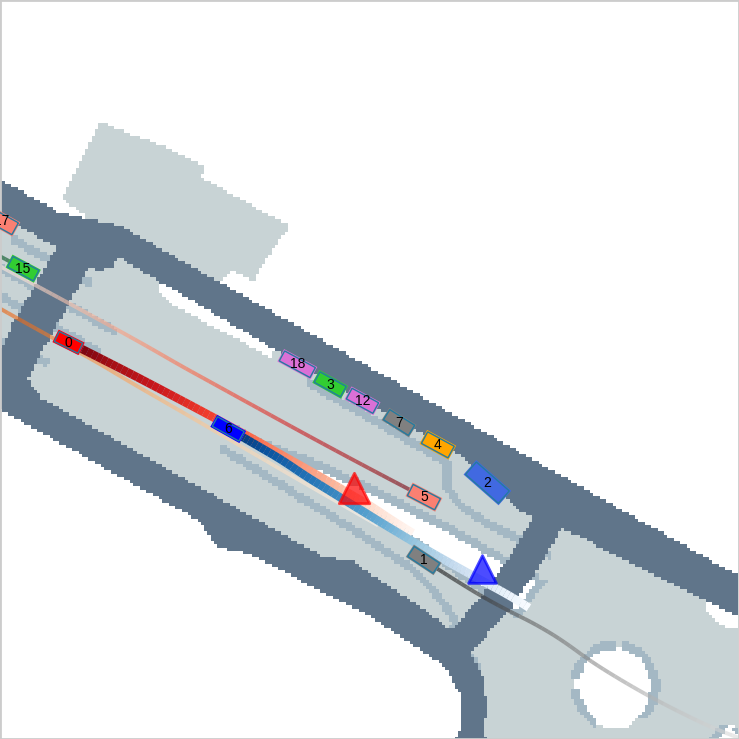}}
\subfloat[][\small{\method collision} \label{fig:gpt_example-ctg++-collision}]{\includegraphics[width=0.24\textwidth]{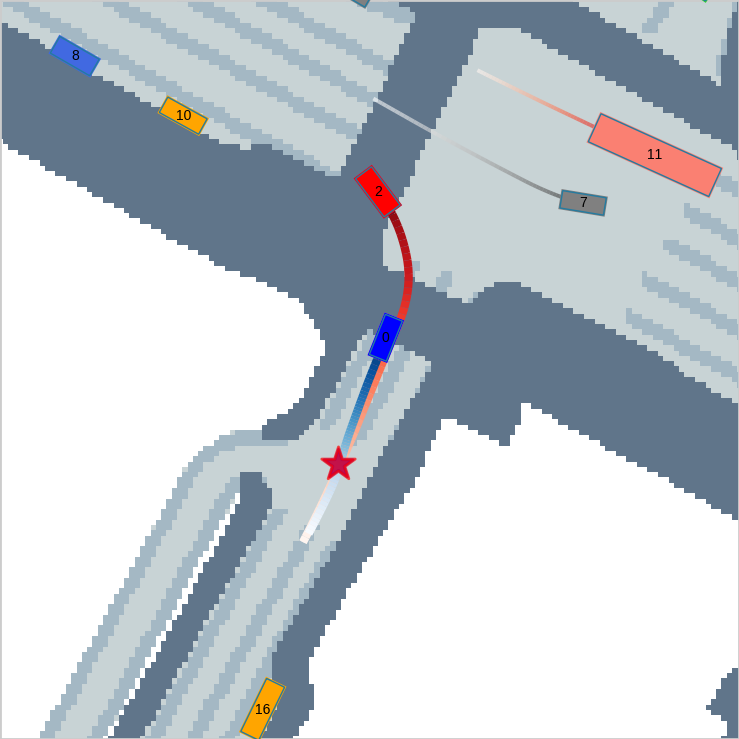}}
\subfloat[][\small{CTG collision} \label{fig:gpt_example-ctg-collision}]{\includegraphics[width=0.24\textwidth]{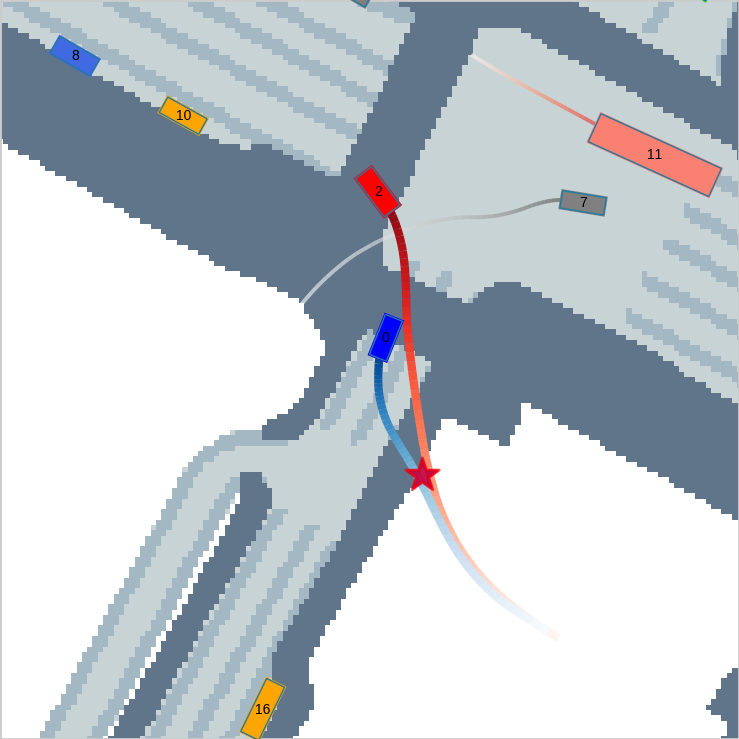}}
\caption{\small{Generated trajectories for query ``\textcolor{red}{vehicle A} should always keep within 10-30m from \textcolor{blue}{vehicle B}" and ``\textcolor{red}{vehicle A} should collide with \textcolor{blue}{vehicle B}", respectively. The collision and offroad locations are marked in $\medstar$ and $\triangle$. For both \method and CTG, our language interface generate query compliant trajectories. However, \method does not sacrifice other aspects like keeping on-road and smoothness while CTG does.}}
\label{fig:gpt_example}
\end{figure}

We conduct two case studies on queries for common traffic behaviors demonstrating \method can effectively generate trajectories satisfying user's queries. Both queries involve the interaction of two vehicles. The first query ``vehicle A should always keep within 10-30m from vehicle B" (\textit{GPT keep distance}) is a common scene in the real world and the generated scene can be used to test vehicle following.  \Cref{fig:gpt_example-ctg++-keepdistance} shows the simulation generated by \method: vehicle A (in \textcolor{red}{red}) follows vehicle B (in \textcolor{blue}{blue}) to go straight and slightly turn right with a safe distance as specified. Both vehicles along with the background vehicles have smooth motion without critical failures during the rollout. 
The second query ``vehicle A should collide with vehicle B" (\textit{GPT collision}) makes it possible to potentially 
generate scenarios like
%reconstruct
those from a crash report \cite{nhtsa_crash_viewer} and test a vehicle's reaction in dangerous situations. \Cref{fig:gpt_example-ctg++-collision} shows the simulation generated by \method: vehicle A collides with vehicle B while turning right, as desired. The motion of all vehicles is smooth and does not have collisions other than the one requested. The generated query-compliant trajectories for both cases show that our proposed language interface in \method enables effective text-to-traffic generation.

% \noindent\textbf{Vehicle cuts in another vehicle.}

\subsection{Evaluation of Traffic Model}
\label{sec:experiment-traffic_model_evaluation}
\begin{table}[ht]
\centering
\scriptsize
\caption{
    \small{Quantitative results of \method and the baselines under GPT-generated rules and STL rules.}\label{table:main_result}
}
\resizebox{1.0\textwidth}{!}{
\begin{tabularx}{1.26\textwidth}{l|cccc|cccc|cccc|cccc}
\toprule
            & \multicolumn{4}{c|}{GPT keep distance}                                                                                     & \multicolumn{4}{c|}{GPT collision}                                                                                         & \multicolumn{4}{c|}{no collision}                                                                                          & \multicolumn{4}{c}{speed limit}                                                                                           \\
            & \multicolumn{1}{l}{fail}     & \multicolumn{1}{l}{rule}     & \multicolumn{1}{l}{real}     & \multicolumn{1}{l|}{rel real}   & \multicolumn{1}{l}{fail}     & \multicolumn{1}{l}{rule}     & \multicolumn{1}{l}{real}     & \multicolumn{1}{l|}{rel real}   & \multicolumn{1}{l}{fail}     & \multicolumn{1}{l}{rule}     & \multicolumn{1}{l}{real}     & \multicolumn{1}{l|}{rel real}   & \multicolumn{1}{l}{fail}     & \multicolumn{1}{l}{rule}     & \multicolumn{1}{l}{real}     & \multicolumn{1}{l}{rel real}   \\
\midrule
BITS & {\color[HTML]{000000} 0.183} & 2.615                        & 0.116                        & 0.362                        & \textbf{0.176} & 0.660                        & 0.107                        & 0.359                        & 0.092                        & 0.065                        & 0.099                        & 0.352                        & {\color[HTML]{000000} 0.111} & 0.559                        & 0.104                        & 0.352                        \\
BITS+opt    & 0.240                        & \textbf{0.000} & 0.097                        & 0.360                        & 0.277                        & 0.130                        & \textbf{0.068} & 0.362                        & {\color[HTML]{000000} 0.109} & {\color[HTML]{000000} 0.041} & 0.070                        & 0.353                        & 0.162                        & 0.120                        & {\color[HTML]{000000} 0.058} & 0.353                        \\
CTG         & 0.343                        & \textbf{0.000} & \textbf{0.077} & {\color[HTML]{000000} 0.342} & 0.356                        & \textbf{0.000} & {\color[HTML]{000000} 0.074} & {\color[HTML]{000000} 0.349} & 0.142                        & 0.052                        & {\color[HTML]{000000} 0.044} & {\color[HTML]{000000} 0.346} & 0.128                        & {\color[HTML]{000000} 0.029} & 0.075                        & {\color[HTML]{000000} 0.350} \\
CTG++       & \textbf{0.173} & \textbf{0.000} & \textbf{0.077} & \textbf{0.331} & {\color[HTML]{000000} 0.264} & \textbf{0.000} & 0.085                        & \textbf{0.331} & \textbf{0.084} & \textbf{0.036} & \textbf{0.040} & \textbf{0.332} & \textbf{0.083} & \textbf{0.028} & \textbf{0.043} & \textbf{0.344} \\
\bottomrule
\end{tabularx}
}
\resizebox{1.0\textwidth}{!}{
\begin{tabularx}{1.4\textwidth}{l|cccc|cccc|ccccc|ccccc}
\toprule
            & \multicolumn{4}{c|}{target speed}                                                                                          & \multicolumn{4}{c|}{no offroad}                                                                                            & \multicolumn{5}{c|}{goal waypoint+target speed}                                                                                                           & \multicolumn{5}{c}{stopregion+offroad}                                                                                                                   \\

            & \multicolumn{1}{l}{fail}     & \multicolumn{1}{l}{rule}     & \multicolumn{1}{l}{real}     & \multicolumn{1}{l|}{rel real}   & \multicolumn{1}{l}{fail}     & \multicolumn{1}{l}{rule}     & \multicolumn{1}{l}{real}     & \multicolumn{1}{l|}{rel real}   & \multicolumn{1}{l}{fail}     & \multicolumn{1}{l}{rule1}    & \multicolumn{1}{l}{rule2}    & \multicolumn{1}{l}{real}     & \multicolumn{1}{l|}{rel real}   & \multicolumn{1}{l}{fail}     & \multicolumn{1}{l}{rule1}    & \multicolumn{1}{l}{rule2}    & \multicolumn{1}{l}{real}     & \multicolumn{1}{l}{rel real}   \\
\midrule
BITS & 0.111                        & 1.526                        & 0.114                        & \textbf{0.355} & \textbf{0.097} & 0.018                        & 0.099                        & 0.355                        & {\color[HTML]{000000} 0.111} & 2.261                        & 1.010                        & 0.115                        & 0.358                        & 0.121                        & 0.005                        & 1.690                        & {\color[HTML]{000000} 0.068} & 0.353                        \\
BITS+opt    & 0.257                        & 0.742                        & \textbf{0.072} & {\color[HTML]{000000} 0.356} & 0.105                        & 0.005                        & 0.100                        & 0.358                        & 0.254                        & 3.681                        & 0.746                        & 0.079                        & {\color[HTML]{000000} 0.342} & {\color[HTML]{000000} 0.095} & 0.020                        & 2.053                        & 0.097                        & 0.354                        \\
CTG         & {\color[HTML]{000000} 0.091} & {\color[HTML]{000000} 0.281} & 0.105                        & 0.379                        & 0.172                        & \textbf{0.002} & {\color[HTML]{000000} 0.042} & {\color[HTML]{000000} 0.346} & 0.118                        & {\color[HTML]{000000} 2.388} & \textbf{0.387} & {\color[HTML]{000000} 0.052} & 0.345                        & 0.128                        & \textbf{0.002} & {\color[HTML]{000000} 0.808} & \textbf{0.040} & {\color[HTML]{000000} 0.336} \\
CTG++       & \textbf{0.060} & \textbf{0.274} & {\color[HTML]{000000} 0.082} & 0.370                        & \textbf{0.097} & {\color[HTML]{000000} 0.004} & \textbf{0.038} & \textbf{0.328} & \textbf{0.101} & \textbf{2.352} & {\color[HTML]{000000} 0.396} & \textbf{0.038} & \textbf{0.338} & \textbf{0.081} & {\color[HTML]{000000} 0.003} & \textbf{0.411} & 0.076                        & \textbf{0.324} \\
\bottomrule
\end{tabularx}
}
\end{table}

We assess the traffic model component of \method under the two GPT-generated rules (as described in \Cref{sec:experiment-gpt_case_study}) and six STL rules from \cite{zhong2022guided} (see \Cref{sec:rules_details} for details of each rule). The quantitative results in \Cref{table:main_result} underscore \method's superiority over baselines with a good balance between stability, rule satisfaction, and realism. Specifically, \method secures the lowest failure rate and scene-level realism deviation in 7 out of 8 settings, reflecting its effective scene-level modeling and enhanced interaction dynamics. Furthermore, \method also achieves the lowest rule violation and agent-level realism deviation for the majority of settings, demonstrating that enhanced interaction modeling does not compromise agent-level realism or rule adherence. 

\begin{wraptable}{L}{0.42\textwidth}
% \vspace{-3mm}
\scriptsize
\centering
\caption{\small{Ablation study of \method features.}}
\label{tab:ablation}
\begin{tabular}{l|cccc}
\toprule
                & fail & rule & real & rel real \\
\midrule
\method           &    \textbf{0.173}	& \textbf{0.000}	& \textbf{0.077}	& \textbf{0.331} \\
% \method spatial every step & 0.211       & 0.000 & 0.073 & 0.331 \\
\method no edge   &    0.227	& \textbf{0.000}	& \textbf{0.077}	& 0.341     \\
\method scene     &    0.886	& 1.043	& 0.127	& 0.392     \\
\bottomrule
\end{tabular}
% \vspace{-3mm}
\end{wraptable}

Qualitatively, \method generates rule-compliant trajectories featuring more realistic motion with fewer instances of collisions or off-road incidents than the baselines. We provide examples of CTG-generated trajectories (when using the same language generated loss functions) for the same scenes as those previously shown for \method (\Cref{fig:gpt_example-ctg-keepdistance} and \Cref{fig:gpt_example-ctg-collision}). Specifically, in \Cref{fig:gpt_example-ctg-keepdistance}, CTG's trajectories display off-road instances (indicated by $\triangle$) involving two vehicles and a background vehicle. Similarly, in \Cref{fig:gpt_example-ctg-collision}, for CTG's trajectories, the collision between vehicles A and B (in $\medstar$) occurs off-road, and the background vehicle has a curvy, unrealistic path. See \Cref{sec:stl_qualtative} for qualitative comparisons under STL rules.

% Qualitatively, \method generates rule-compliant trajectories featuring more realistic motion with fewer instances of collisions or off-road incidents than the baselines. As illustrated in \Cref{sec:experiment-gpt_case_study}, \method-generated trajectories in \Cref{fig:gpt_example-ctg++-keepdistance} satisfy user queries, exhibiting smooth, realistic motion devoid of failures. By contrast, CTG's trajectories (created using our proposed language interface for a fair comparison) shown in \Cref{fig:gpt_example-ctg-keepdistance} include off-road incidents (in $\triangle$) involving two vehicles and a background vehicle. Similar findings are observed in \Cref{fig:gpt_example-ctg++-collision}, where \method-generated trajectories meet user-defined criteria, including a collision between two vehicles A and B, while maintaining smooth motion without additional collisions or off-road incidents. In contrast, CTG's trajectories (\Cref{fig:gpt_example-ctg-collision}) result in a collision between vehicles A and B (in $\medstar$) occurring off-road, and the background vehicle has a curvy path. 
% These examples show the superiority of \method over CTG as it satisfies the queries without sacrificing much other aspects (e.g., stability and realism). 

\begin{wrapfigure}{R}{0.5\textwidth}
\centering
\subfloat[][\small{\method}]{\includegraphics[width=0.24\textwidth]{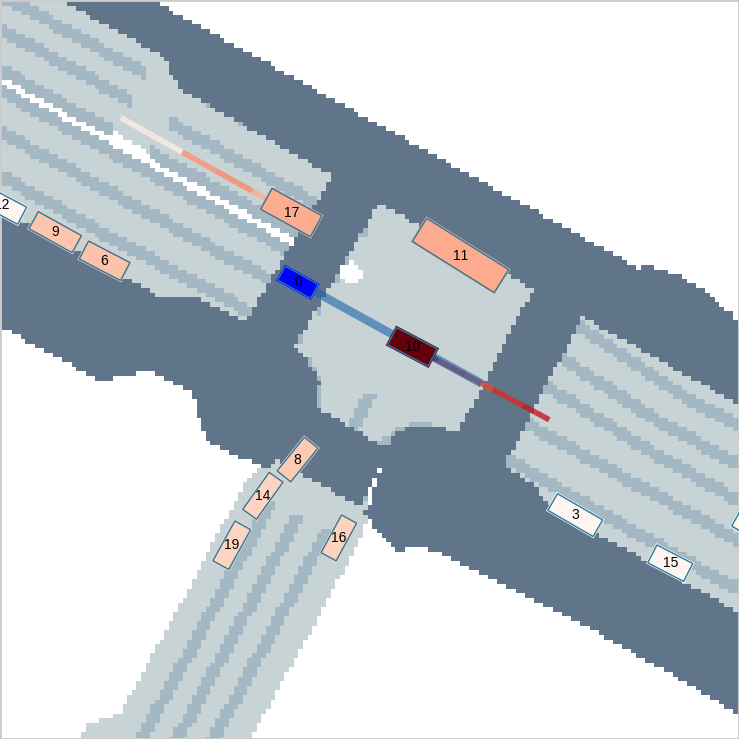}}
\subfloat[][\small{\method no edge}]{\includegraphics[width=0.24\textwidth]{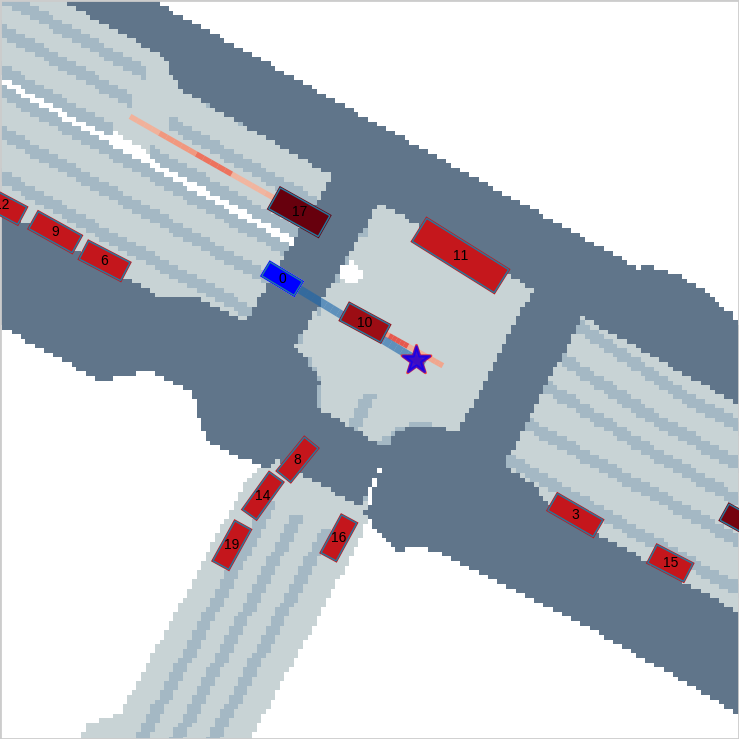}}

\caption{\small{Darker red means higher attention by the \textcolor{blue}{blue vehicle}. Without edge information, \method no edge results in a collision (in $\medstar$).}}\label{fig:attn_example}
\end{wrapfigure}

% 2 examples of CTG++ VS CTG VS BITS

% \noindent\textbf{Conditioning and Counterfactual Analyses.}
% % 2 examples of counter-factual behaviors

\noindent\textbf{Ablation Study.}
We evaluate the efficacy of our spatial module and the utilization of agent coordinates. As demonstrated in \Cref{tab:ablation}, \method surpasses \textbf{CTG++ no edge} (i.e., \method but having $\mathbf{e}_t^{ij}$ in \Cref{eq:edge} replaced by zeros) 
% without edge information (\textbf{CTG++ no edge}) 
% for which we replace the edge information with zeros 
and \textbf{CTG++ scene} (i.e., \method but using scene-centric coordinates) under the \textit{GPT Keep Distance} rule. The absence of edge information leads to increased failure rates due to decreased awareness of interactions. Moreover, the use of scene-centric coordinates notably inflates failure rates and realism deviations, as the traffic rapidly deviates from the training distribution during the rollout.
To visualize the effectiveness of spatial attention, we display the attention maps for a vehicle of interest (highlighted in \textcolor{blue}{blue}) for both \method and \textbf{CTG++ no edge} when guidance is not applied. As depicted in \Cref{fig:attn_example}, \method guides the vehicle to pay attention to relevant neighboring agents, while \textbf{CTG++ no edge} results in arbitrary attention and a consequent collision (marked as $\medstar$) with the vehicle ahead.

% prev collision/offroad loss, rasterized map, no edge, scene-centric coordinate

% attention map visualization

% Using rasterized map also causes higher failure rate due to less fine-grained interaction between agents and map.
\section{Conclusion}

\noindent\textbf{Summary.} In this paper, we present, \method, a language-guided scene-level conditional diffusion model for realistic, query-compliant traffic simulation. In particular, we leverage an LLM for translating a user query into a differentiable loss function and propose a scene-level conditional diffusion model (with a spatial-temporal transformer architecture) to translate the loss function into realistic, query compliant trajectories. Extensive evaluation demonstrates the effectiveness of our approach.

\noindent\textbf{Limitations and Future Work.} 
\method currently does not support complex commands involving many interactions with map (see \Cref{sec:gpt-failure}). Our framework can be extended by using a multi-modal LLM that takes in vision data (e.g., bird's-eye view map) for a finer control of the traffic and thus supports more complex commands. Second, our framework does not support automatic error detection and fixing for the GPT4 generated loss function. As the loss function (in code format) can have wrong semantics, it is necessary to instruct GPT4 to detect and repair it. Our framework can be potentially extended to provide the simulation running results to GPT4 to iteratively instruct it to fix the generated loss function. Third, current trajectory generation is relatively slow and take about 1 minute to generate each simulated scenario. Recent work which uses distillation \cite{salimans2022progressive} to greatly speed up the generation process can be leveraged to reduce the time cost.
Our work opens up many possibilities including adapting our architecture to general multi-agent robotic tasks and using our proposed two-step approach for other tasks with no paired text-behavior data.

\clearpage

\section*{Acknowledgment}
The authors want to thank Sushant Veer and Shuhan Tan for valuable discussions. This work started when Ziyuan Zhong interned at NVIDIA Research. He is also supported by NSF CCF 1845893 and IIS 2221943.

%===============================================================================

% The acknowledgments are automatically included only in the final and preprint versions of the paper.
% \acknowledgments{If a paper is accepted, the final camera-ready version will (and probably should) include acknowledgments. All acknowledgments go at the end of the paper, including thanks to reviewers who gave useful comments, to colleagues who contributed to the ideas, and to funding agencies and corporate sponsors that provided financial support.}

%===============================================================================

\bibliography{references}
\clearpage
\appendix
\renewcommand{\thefigure}{A\arabic{figure}}
\setcounter{figure}{0}

\section{Algorithm of Training and Sampling in Details}
We mostly follow the training and sampling procedures from \cite{zhong2022guided} and show the detailed algorithms for training and sampling in the following.

\subsection{Training}
\label{sec:algo-training}

{\centering
\begin{figure}[ht]
\vspace{-2mm}
% \scalebox{0.73}{
\begin{minipage}{\linewidth}
  \begin{algorithm}[H]
  \scriptsize
    \caption{Training}
    \label{alg:training}
    \begin{algorithmic}[1]
    \STATE \textbf{Require} a real-world driving dataset $D$, conditional diffusion model to train $\boldsymbol{\mu}^0_\theta$, transition function $f$, denoising steps $K$. \\
    \WHILE{not converge}
        \STATE $\mathbf{c}, \btau{0} \sim D$
        \STATE $k \sim \{1,...,K\}$
        \STATE $\epsilon \sim \mathcal{N}(\bm{0}, \bm{I})$

        \STATE Corrupt action trajectory $\btau{k}_a = \sqrt{\bar{\alpha}_{k}} \btau{0}_a + \sqrt{1-\bar{\alpha}_{k}} \epsilon $ with $\bar{\alpha}_{k} = \prod_{l=0}^k 1 - \beta_l$

        \STATE Get the corresponding state trajectory $\btau{k}_s = f(s_0, \btau{k}_a)$
        
        \STATE Use model to predict the uncorrupted trajectory $\hat{\boldsymbol{\tau}}^0_a = \boldsymbol{\mu}^0_\theta(\btau{k},k,\bc{})$
        \STATE Get the predicted state trajectory $\hat{\boldsymbol{\tau}}^0 = [ \hat{\boldsymbol{\tau}}^0_a; \; f(s_0, \hat{\boldsymbol{\tau}}^0_a)]$
        \STATE Take gradient step on $\nabla_\theta || \btau{0} - \hat{\boldsymbol{\tau}}^0 ||^2$
    \ENDWHILE
    \end{algorithmic}
  \end{algorithm}
\end{minipage}
\end{figure}
}

Contrary to \cite{zhong2022guided}, which samples trajectories at the agent level, we opt for scene-level trajectory sampling, allowing the model to make joint predictions on all scene agents. The process is detailed in \Cref{alg:training}. During each training iteration, the context $\mathbf{c}$ and the ground-truth trajectory $\btau{0}$ are sampled from a real-world driving dataset, and the denoising step $k$ is uniformly selected from $\{1, \dots, K\}$.
We derive the noisy input $\btau{k}$ from $\btau{0}$ by initially corrupting the action trajectory via $\btau{k}_a = \sqrt{\bar{\alpha}_{k}} \btau{0}_a + \sqrt{1-\bar{\alpha}_{k}} \epsilon $, with $\epsilon \sim \mathcal{N}(0,\mathbf{I})$ and $\bar{\alpha}_{k} = \prod_{l=0}^k 1 - \beta_l$. Subsequently, the corresponding state is computed as $\btau{k}_s = f(s_0, \btau{k}_a)$.
The diffusion model indirectly parameterizes $\boldsymbol{\mu}_\theta$ in \cref{eq: denoise_p} by predicting the uncorrupted trajectory $\hat{\boldsymbol{\tau}}^0 = [ \hat{\boldsymbol{\tau}}^0_a; f(s_0, \hat{\boldsymbol{\tau}}^0_a)]$, where $\hat{\boldsymbol{\tau}}^0_a = \boldsymbol{\mu}^0_\theta(\btau{k},k,\bc{})$ is the network's direct output (see \cite{li2022nlpdiffusion,janner2022diffuser,nichol2021improved}). We then use a simplified loss function to train the model as follows:
\begin{equation}
\begin{aligned}
\label{eq: loss_simple}
L(\theta) = \mathbb{E}_{\epsilon,k, \btau{0}, \bc} \left[ || \btau{0} - \hat{\boldsymbol{\tau}}^0 ||^2 \right].
% L(\theta, \psi) = \mathbb{E}{\epsilon,\btau{0},k, \bc} \Vert \epsilon - \epsilon_{(\theta, \psi)}(\btau{k}, k, \bc) \Vert,
\end{aligned}
\end{equation}
A cosine variance schedule~\cite{janner2022diffuser,nichol2021improved} is utilized in the diffusion process, employing $K=100$ diffusion steps.

\subsection{Sampling}
\label{sec:algo-sampling}
We show the guided sampling algorithm in \Cref{alg:sampling} which is directly from \cite{zhong2022guided} as the notations and procedure remain the same. The key difference is that our diffusion model formulation and backbone models are all at scene-level rather than agent-level as in \cite{zhong2022guided}. The scene-level formulation helps to improve scene-level realism and decrease failure rates as the agents' interactions can be captured by the model inherently.

{\centering
\begin{figure}[ht]

% \scalebox{0.73}{
\begin{minipage}{\linewidth}
  \begin{algorithm}[H]
  \scriptsize
    \caption{Guided Sampling}
    \label{alg:sampling}
    \begin{algorithmic}[1]
    \STATE \textbf{Require} conditional diffusion model $\boldsymbol{\mu}_\theta$, transition function $f$, guide $\mathcal{J}$, scale $\alpha$, covariances $\boldsymbol{\Sigma}^k$, diffusion steps $K$, inner gradient descent steps $W$, number of actions to take before re-planning $l$. \\
        
    \WHILE{not done}
        \STATE Observe state $s_0$ and context $\bc$
        \STATE Initialize trajectory $\btau{K}_a \sim \mathcal{N}(\bm{0}, \bm{I})$; $\btau{K}_s = f(s_0, \btau{K}_a)$; $\btau{K}=[\btau{K}_a; \btau{K}_s]$
        \FOR{$k = K, \ldots, 1$}
            \STATE $\bmu := \hat{\btau{}}_a^{k-1} = \bmu_\theta(\btau{k}, k, \bc)$ \\
            \STATE $\bmu^{(0)} = \bmu$\\
            \FOR{$j = 1, \ldots, W$}
                \STATE $\bmu^{(j)} = \bmu^{(j-1)} + \alpha \nabla \mathcal{J}(\bmu^{(j-1)})$\\
                \STATE $\Delta \bmu = |\bmu^{(j)}-\bmu^{(0)}|$
                \STATE $\Delta \bmu \gets \textrm{clip}(\Delta \bmu, -\beta_k, \beta_k)$\\
                \STATE $\bmu^{(j)} \gets \bmu^{(0)} + \Delta \bmu$ \\
            \ENDFOR
            \STATE $\btau{k-1}_a \sim \mathcal{N}(\bmu^{(M)}, \bSigma^k)$; $\btau{k-1}_s = f(s_0, \btau{k-1}_a)$;\\ $\btau{k-1}=[\btau{k-1}_a; \btau{k-1}_s]$\\
        \ENDFOR
        \STATE Execute first $l$ actions of trajectory $\btau{0}_{a}$
    \ENDWHILE
    \end{algorithmic}
  \end{algorithm}
\end{minipage}
\end{figure}
}

Following \cite{zhong2022guided, li2022nlpdiffusion}, the predicted mean is a weighted sum between the predicted clean action trajectory and the input action trajectory from last denoising step:
\begin{equation}
    \hat{\btau{}}_a^{k-1} = \bmu_\theta(\btau{k}, k, \bc) = \frac{\sqrt{\bar{\alpha}_{k-1}} \beta_k}{1-\bar{\alpha}_k} \hat{\btau{}}^{0}_a+\frac{\sqrt{\alpha_k}\left(1-\bar{\alpha}_{k-1}\right)}{1-\bar{\alpha}_k} \btau{k}_a
\end{equation}
The process of perturbing the predicted means from the diffusion model using gradients of a specified objective is summarized in \cref{alg:sampling}. 
Following \cite{zhong2022guided}, we use an iterative projected gradient descent with the Adam optimizer 
%which gives flexibility to trade off rule compliance and realism by adjusting learning rate and the number of optimization steps.
and \textit{filtration}, i.e., we guide several samples from the diffusion model and choose the one with the best rule satisfaction based on $\mathcal{J}$.

\section{More Details on Architecture}
\subsection{Detailed Architecture}
\begin{figure}[ht!]
\centering
\includegraphics[width=\textwidth]{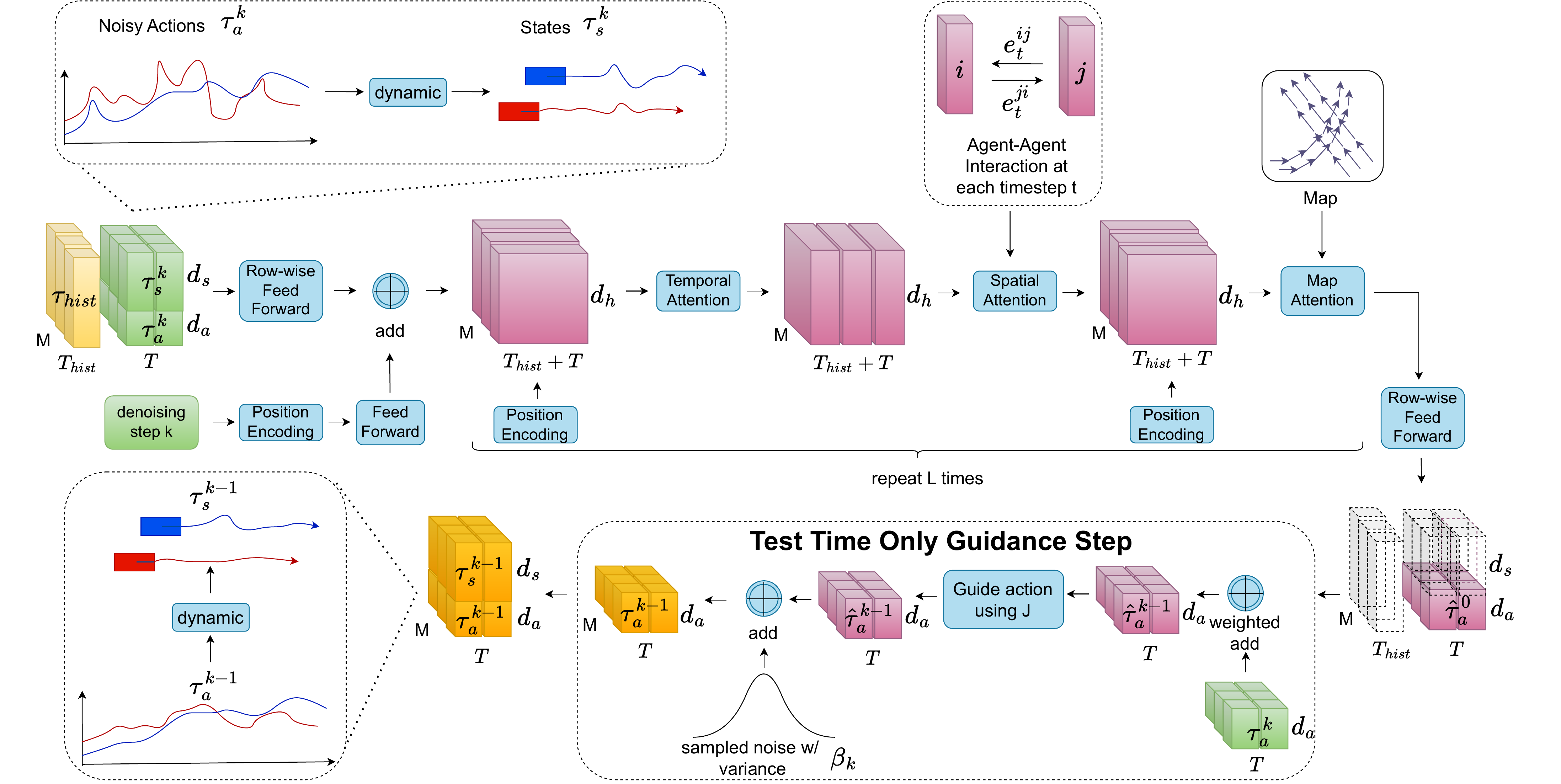}
\caption{\small{Test time denoising step using scene-level spatial-temporal transformer. $d_s$, $d_a$, and $d_h$ represent the dimensions of action, state, and latent for each vehicle per timestep.}}
\label{fig:state_action_detailed}
\end{figure}

We show the detailed data flow of our proposed architecture in \Cref{fig:state_action_detailed}. Its main difference with the simplified architecture shown in \Cref{fig:state_action} is that we show position encoding, rFFN, and the details of the guidance module explicitly. 

\subsection{Gated Attention}
\label{sec:gated_attention}
Following \cite{Zhou_2022_CVPR}, we use a variant of the original scaled dot-product attention block. In particular, we use a gating function to fuse the environmental features $m^t_i$ with the central agent’s features $h^t_i$, enabling the block to have more control over the feature update. The resulting query, key, and value vectors of the social attention layer are taken as inputs to the block:
\begin{equation}
\begin{gathered}
\boldsymbol{\alpha}^i_t=\operatorname{softmax}\left(\frac{\mathbf{q}^{i^{\top}}_{t}}{\sqrt{d_k}} \cdot\left[\left\{\mathbf{k}^{i j}_t\right\}_{j \in \mathcal{N}_i}\right]\right), \\
\mathbf{m}^i_t=\sum_{j \in \mathcal{N}_i} \boldsymbol{\alpha}^{i j}_t \mathbf{v}^{i j}_t, \\
\mathbf{g}^i_t=\operatorname{sigmoid}\left(\mathbf{W}^{\text {gate }}\left[\mathbf{h}^i_t, \mathbf{m}^i_t\right]\right), \\
\hat{\mathbf{h}}^i_t=\mathbf{g}^i_t \odot \mathbf{W}^{\text {self }} \mathbf{h}^i_t+\left(1-\mathbf{g}^i_t\right) \odot \mathbf{m}^i_t,
\end{gathered}
\end{equation}
where $\mathcal{N}_i$ is the set of agent i’s neighbors (all the agents except the agent itself within a certain social radius), $\mathbf{W}^{\text {gate}}$ and $\mathbf{W}^{\text {self}}$
are learnable matrices, and denotes element-wise product $\odot$.

\section{Qualitative Comparison under STL rules}
\label{sec:stl_qualtative}
\begin{figure}[ht]
\centering
\subfloat[][\small{\method speed limit (0.037)} \label{fig:rules_examples-ctg++_speedlimit}]{\includegraphics[width=0.48\textwidth]{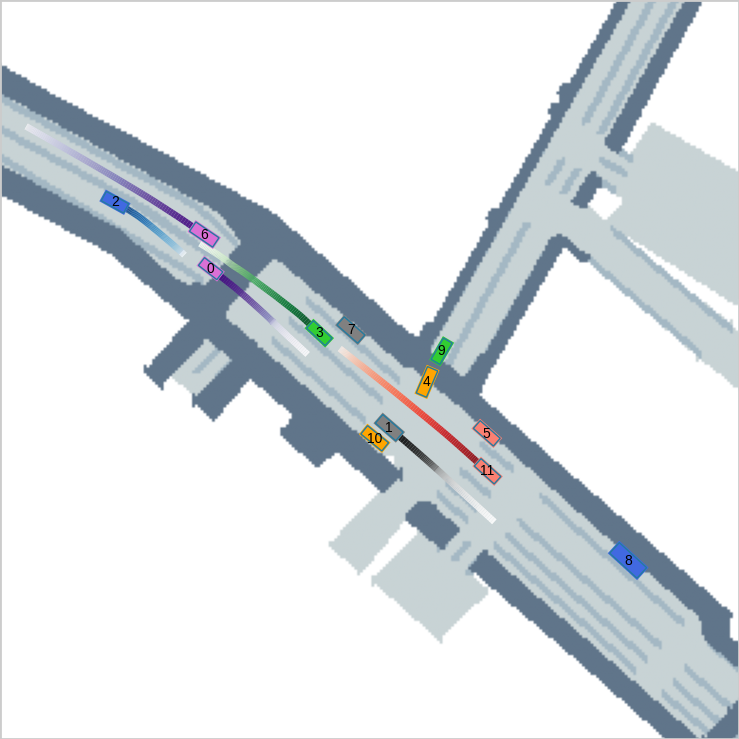}}
\subfloat[][\small{CTG speed limit (0.041)} \label{fig:rules_examples-ctg_speedlimit}]{\includegraphics[width=0.48\textwidth]{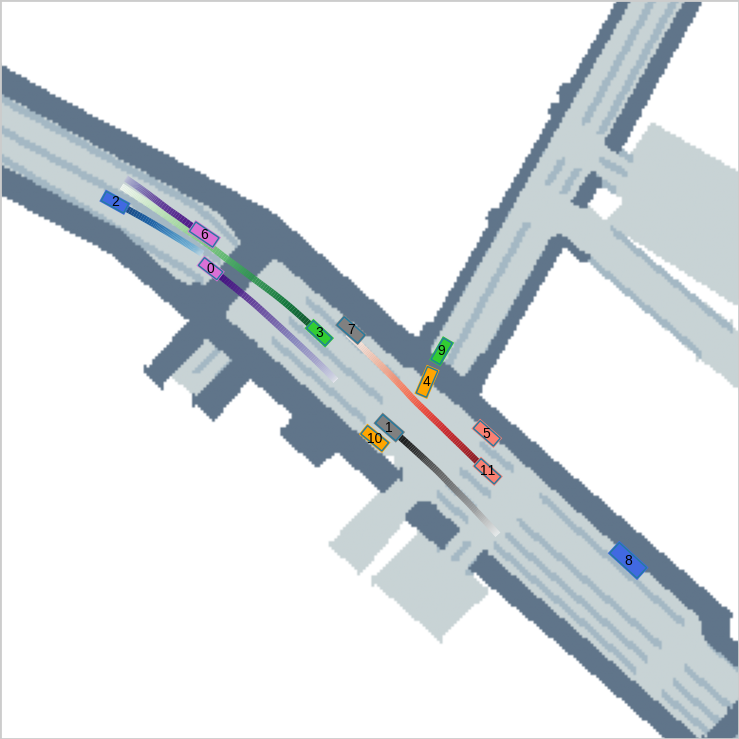}}
\caption{\small{Qualitative comparison between \method and CTG under speed limit STL rule  (the numbers in parentheses represent rule violations). \method achieves lower rule violation than CTG. Besides, CTG involves collision between the blue vehicle and the green vehicle.}}
\label{fig:rules_examples_speedlimit}
\end{figure}

\begin{figure}[ht]
\centering
\subfloat[][\small{\method target speed (0.213)} \label{fig:rules_examples-ctg++_targetspeed}]{\includegraphics[width=0.48\textwidth]{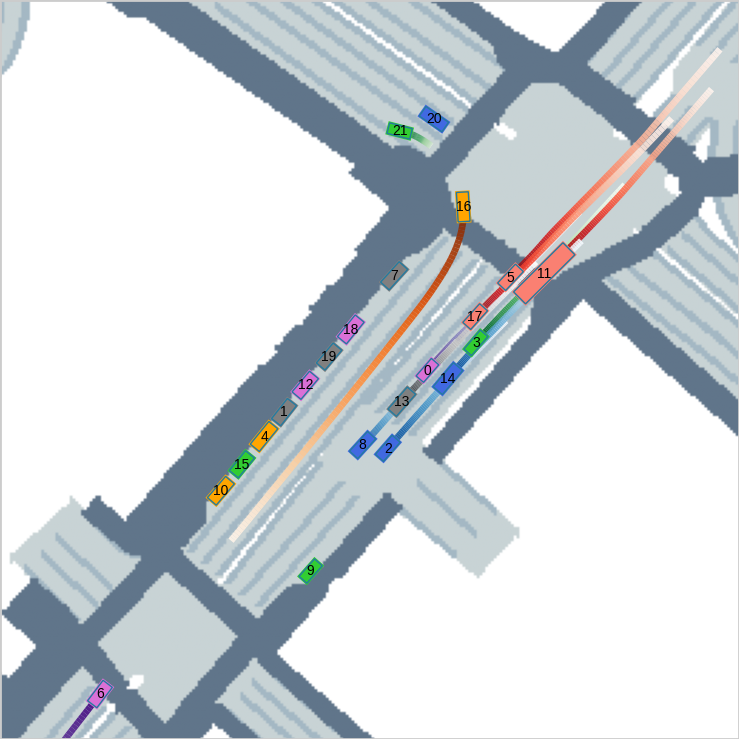}}
\subfloat[][\small{CTG target speed (0.163)} \label{fig:rules_examples-ctg_targetspeed}]{\includegraphics[width=0.48\textwidth]{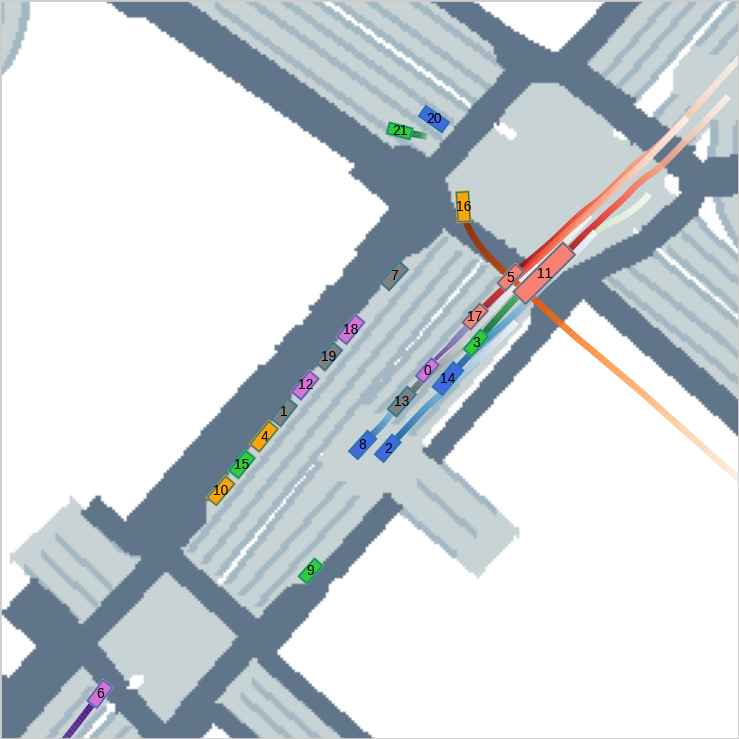}}
\caption{\small{Qualitative comparison between \method and CTG under target speed STL rule  (the numbers in parentheses represent rule violations). Although CTG achieves a bit better target speed rule satisfaction, it involves a vehicle collides with crossing vehicles and then goes off-road.}}
\label{fig:rules_examples_targetspeed}
\end{figure}

\begin{figure}[ht]
\centering
\subfloat[][\small{\method no collision (0)} \label{fig:rules_examples-ctg++_collision}]{\includegraphics[width=0.48\textwidth]{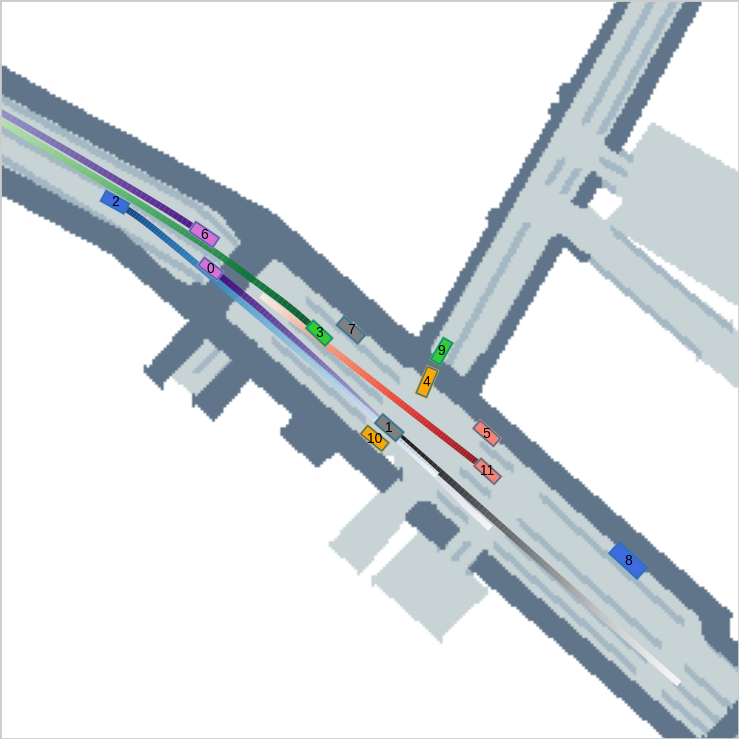}}
\subfloat[][\small{BITS+opt no collision (0)} \label{fig:rules_examples-bits5_collision}]{\includegraphics[width=0.48\textwidth]{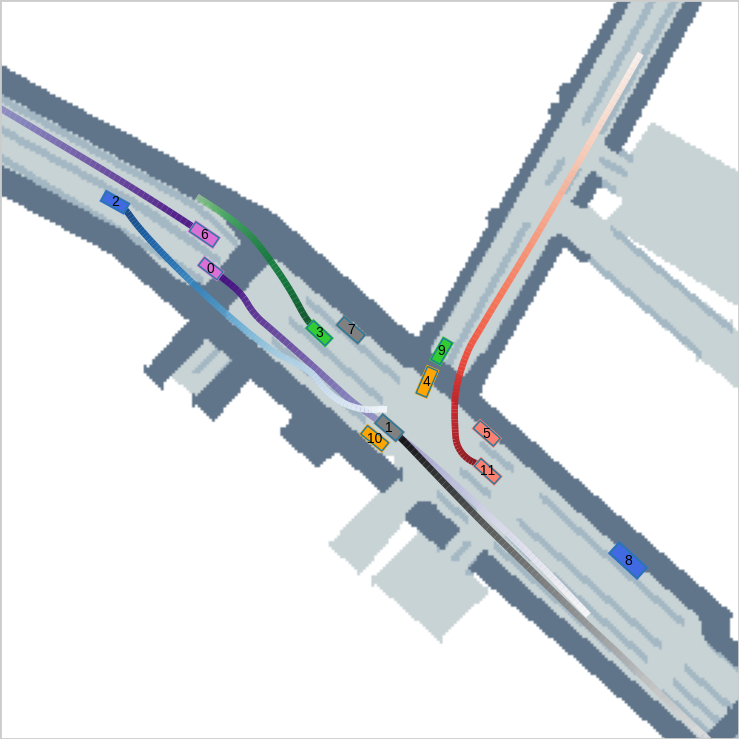}}
\caption{\small{Qualitative comparison between \method and BITS+opt under no collision STL rule  (the numbers in parentheses represent rule violations). Both methods satisfies the rule perfectly as no collision happens. However, BITS+opt have highly curvy, unrealistic trajectories as the cost of satisfying the rule.}}
\label{fig:rules_examples_collision}
\end{figure}

\begin{figure}[ht]
\centering
\subfloat[][\small{\method no off-road (0)} \label{fig:rules_examples-ctg++_offroad}]{\includegraphics[width=0.48\textwidth]{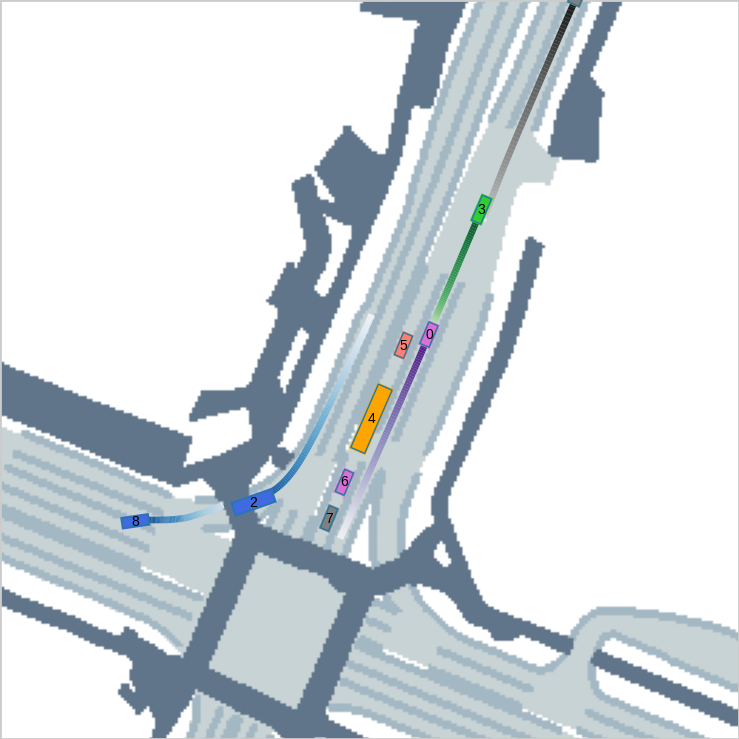}}
\subfloat[][\small{CTG no off-road (0)} \label{fig:rules_examples-ctg_offroad}]{\includegraphics[width=0.48\textwidth]{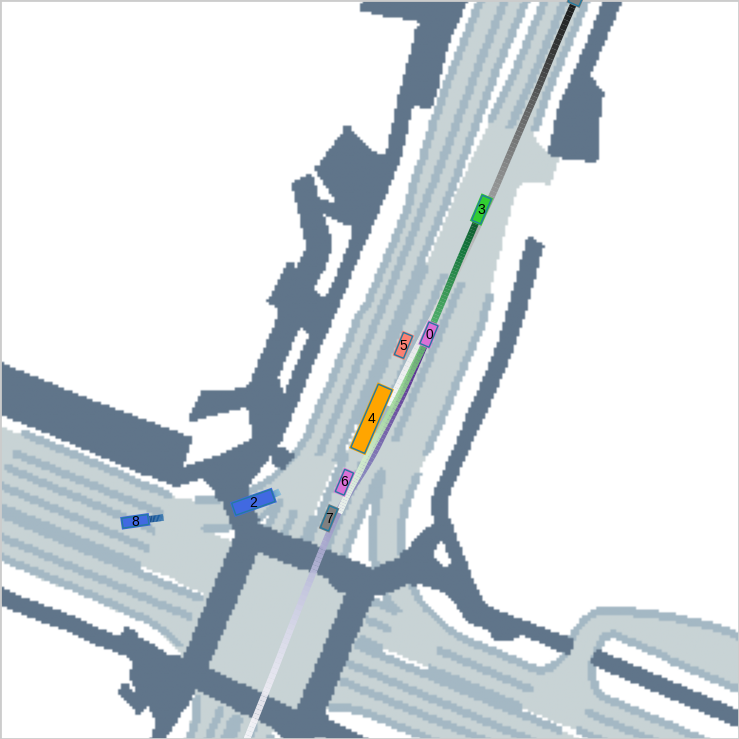}}
\caption{\small{Qualitative comparison between \method and CTG under no off-road STL rule  (the numbers in parentheses represent rule violations). Both methods satisfies the rule perfectly as no off-road happens. However, CTG lead to multiple collisions among the pink vehicle and vehicles that are stationary.}}
\label{fig:rules_examples_offroad}
\end{figure}

\begin{figure}[ht]
\centering
\subfloat[][\small{\method stop sign + no off-road (0, 0)} \label{fig:rules_examples-ctg++_stopsign_offroad}]{\includegraphics[width=0.48\textwidth]{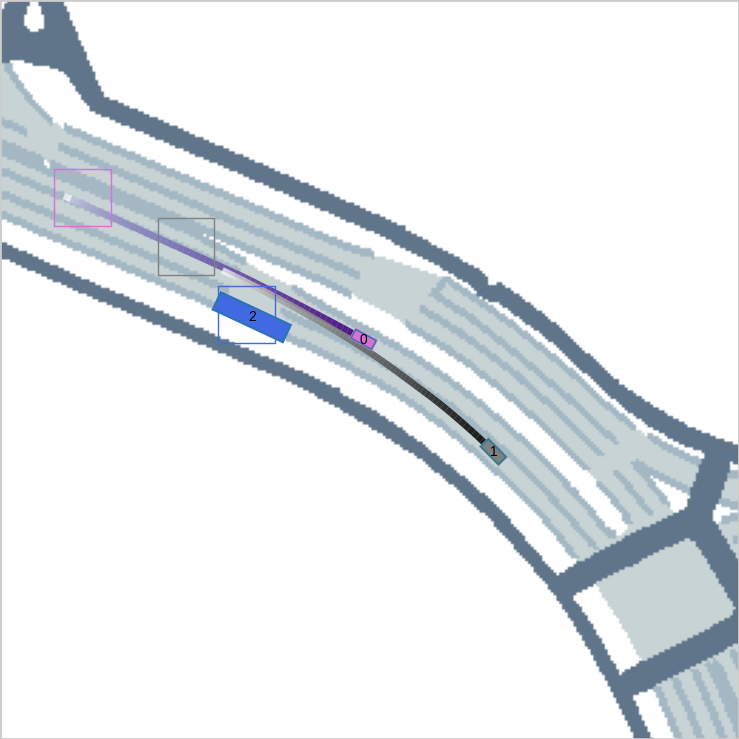}}
\subfloat[][\small{CTG stop sign + no off-road (0.732, 0)} \label{fig:rules_examples-ctg_stopsign_offroad}]{\includegraphics[width=0.48\textwidth]{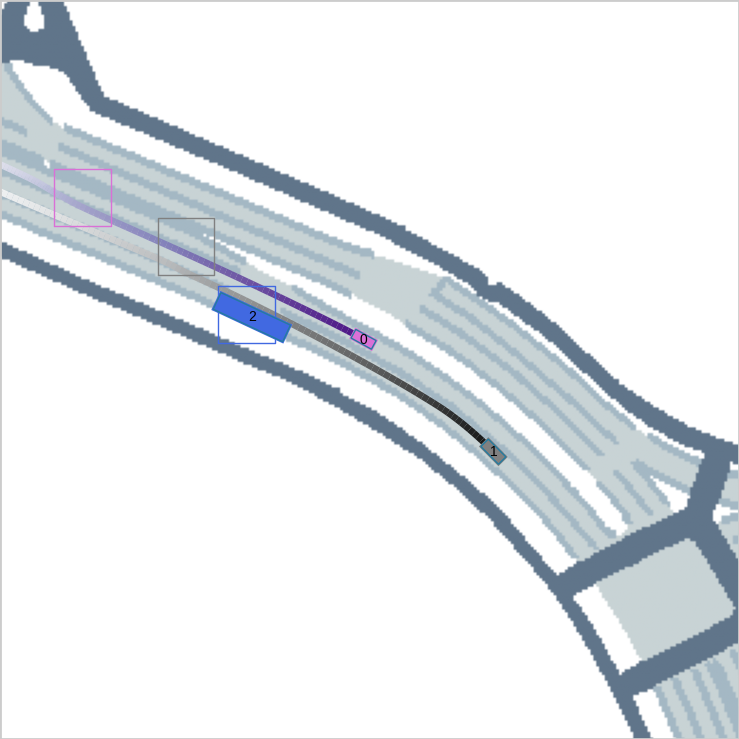}}
\caption{\small{Qualitative comparison between \method and CTG under stop sign and no off-road STL rule (the numbers in parentheses represent rules violations). Vehicles are supposed to stop within the marked bounding boxes without going off-road. \method satisfies both rules while CTG only satisfies the no off-road rule. Besides, CTG involves a collision between the grey vehicle and the blue vehicle.}}
\label{fig:rules_examples_stopsign_offroad}
\end{figure}

\begin{figure}[ht]
\centering
\subfloat[][\small{\method goal waypoint + target speed (0.991, 0.296)} \label{fig:rules_examples-ctg++_goalwaypoint_targetspeed}]{\includegraphics[width=0.48\textwidth]{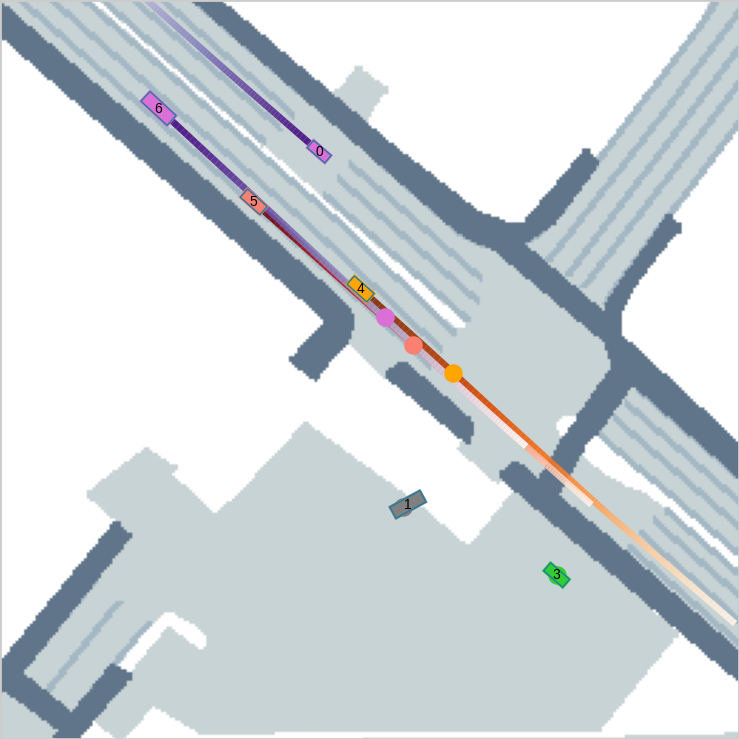}}
\subfloat[][\small{CTG goal waypoint + target speed  (1.24, 0)} \label{fig:rules_examples-ctg_goalwaypoint_targetspeed}]{\includegraphics[width=0.48\textwidth]{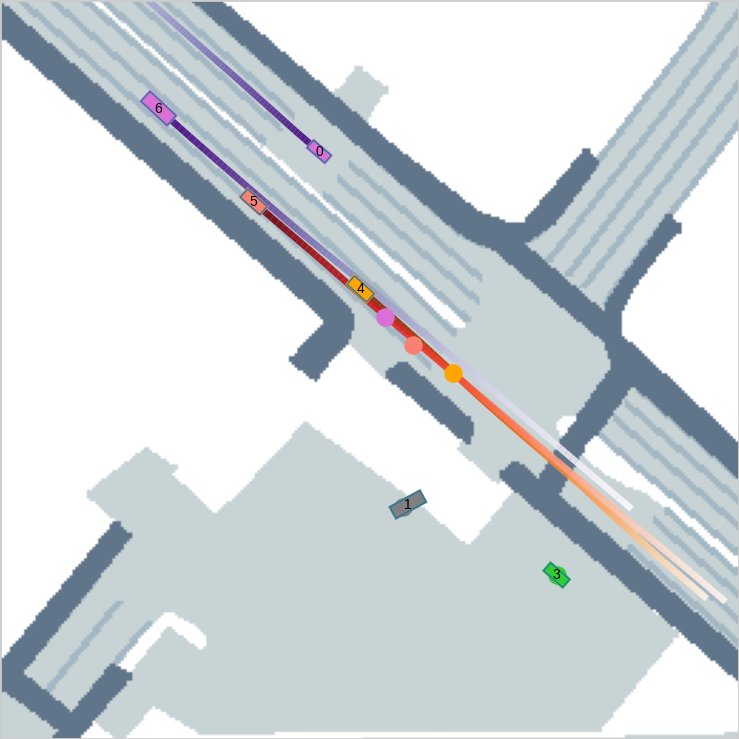}}
\caption{\small{Qualitative comparison between \method and CTG under goal waypoint + target speed STL rule (the numbers in parentheses represent rules violations). Vehicles are supposed to reach the marked waypoints with target speed (same speed as in the dataset). \method satisfies both rules better than CTG. Besides, CTG involves a collision between the two orange vehicles in the end.}}
\label{fig:rules_examples_goalwaypoint_targetspeed}
\end{figure}

In this section, we show a few qualitative examples (\Cref{fig:rules_examples_speedlimit} - \Cref{fig:rules_examples_goalwaypoint_targetspeed}) comparing \method and the strongest baseline (in terms of rule satisfaction) under the STL rules. Overall, \method generates realistic, rule-satisfying trajectories. The baseline method can usually also satisfy the rule. However, their trajectories usually sacrifice one or more of the following aspects: (1) the trajectories are curvy, unrealistic, (2) the trajectories involve off-road accidents, and (3) the agent interaction is sub-optimal leading to collision(s). 

\section{Hyperparameters}
\subsection{Training Hyperparameters}
\method is trained on a machine with Intel i9 12900 and NVIDIA GeForce RTX 3090. It takes approximately 10 hours to train \method for 50K iterations. We use Adam optimizer with a learning rate of 1e-4.

\subsection{Pair Selection Criteria for GPT query based rules}
\label{sec:hyperparameter-pair_selection}
We choose two vehicles A and B in each scene such that they satisfy the following criteria:
\begin{itemize}
    \item Both have current speed larger than 2m/s.
    \item At 0s and 2s, the distance between A and B is within the range 10m to 30m.
    \item At 0s and 2s, the orientation difference between a and b is smaller than 108 degrees (for GPT collision) and 36 degrees (for GPT keep distance).
\end{itemize}
The criteria is a coarse-grained filtration for those pairs that are more likely to have keep distance / collision interactions in the original training dataset.
If more than one pair in the scene satisfy the following criteria, we select the pair with smallest distance. If none of pairs in a scene satisfy the following criteria, we skip the scene. After the filtrations, out of the 100 validation scenes, we have 50 scenes remained for \textbf{GPT collision} and 40 scenes for 
\textbf{GPT keep distance}.

\section{Experiment Details}
\subsection{Quantitative Rules}
\label{sec:rules_details}
We assess the traffic model component of \method under both GPT-generated rules and STL rules from \cite{zhong2022guided}: 
\noindent(1) \textit{GPT Keep Distance} and
\noindent(2) \textit{GPT Collision}, as described in \Cref{sec:experiment-gpt_case_study} (refer to \Cref{sec:hyperparameter-pair_selection} for pair selection);
\noindent(3) \textit{No Collision} dictates that vehicles must avoid collisions with each other;
\noindent(4) \textit{Speed Limit} ensures vehicles do not exceed a set speed limit threshold (75\% quantile of all moving vehicles in a given scene);
\noindent(5) \textit{Target Speed} requires vehicles to maintain a specified speed (50\% of their speed in the ground truth scene) at each time step;
\noindent(6) \textit{No Offroad} prohibits vehicles from leaving the drivable area;
\noindent(7) \textit{Goal Waypoint and Target Speed} instructs vehicles to reach their designated goal while adhering to the specified target speeds;
\noindent(8) \textit{Stop Sign and No Off-road} requires vehicles to halt upon entering a stop sign region and avoid straying off the road.

\subsection{Metrics of Rule Violation}
\label{sec:experiment-metrics_rule}
We provide the details for the metrics we use for measuring rule violation in this section. For all the metrics of rule violation, we average the metrics over all validation scenes. Besides, they are designed such that the smaller the better (i.e., rules are better satisfied).

\noindent\textbf{GPT Keep Distance.} the following vehicle's (in the chosen pair) average L2 distance deviation from the specified range.

\noindent\textbf{GPT Collision.} if a collision happens between the two vehicles in the chosen pair.

\noindent\textbf{No Collision.} collision rate of all vehicles in a scene.

\noindent\textbf{Speed Limit.} average deviation from the speed limit of all vehicles in a scene. 

\noindent\textbf{Target Speed.} average deviation from the target speed of all vehicles in a scene.  

\noindent\textbf{No Offroad.} off-road rate of all vehicles in a scene. We consider a vehicle going off-road if its center goes off-road.

\noindent\textbf{Goal Waypoint.} average vehicle's smallest l2 distance deviation from the specified corresponding goal waypoints of all vehicles in a scene. 

\noindent\textbf{Stop Sign.} average smallest speed within the stop sign region of all vehicles in a scene.

\section{Details of Language Interface}
In this section, we provide more details and limitation of our proposed language interface for traffic simulation.
\subsection{Details of Vehicle Indexing}
In practice, instead of using color for vehicles which is used in \Cref{fig:overview} for better illustration purpose, we use indices according to the context from the driving dataset. The user can tell GPT4 the vehicles to control via their indices, e.g., "vehicle 1 should collide with vehicle 2".

\subsection{Details of Prompting}
\label{sec:gpt-prompt}
In \Cref{fig:gpt_example}, we provide an example of a pre-defined API function and a query-loss function pair. In our experiments, we additionally provide the following API functions:

\noindent\textbf{transform\_coord\_world\_to\_agent\_i.}  this function transform the predicted position and yaw from world coordinate to the agent i coordinate.

\noindent\textbf{select\_agent\_ind.} this function returns the slice of x with index i.

\noindent\textbf{get\_current\_lane\_projection.} this function returns the projection of each vehicle predicted trajectory on its current lane in agent-centric coordinate.

\noindent\textbf{get\_left\_lane\_projection.} this function is similar to get\_current\_lane except it returns the left lane waypoints. If there is no left lane, the original trajectory will be returned.

\noindent\textbf{get\_right\_lane\_projection.} this function is similar to get\_current\_lane except it returns the right lane waypoints. If there is no right lane, the original trajectory will be returned.

In addition to the acceleration loss paired example shown in \Cref{fig:gpt_example}, we provide another query-loss function pair example. "Generate a loss class such that, vehicle 1 should always stay on the left side of vehicle 2." The corresponding function penalize the cases when vehicle 1 not on the left size of vehicle 2. This function provides GPT4 a sense of the relationship between direction and the trajectories.

We additionally specify the dimension of the input trajectory and the input and output of the expected loss function wrapped in a loss class such that GPT4 know which dimension of the trajectory to operate on when needed: "The generated loss class should have a function: forward(x, data\_batch, agt\_mask). x is a tensor representing the current trajectory with shape (B, N, T, 6) where B is the number of vehicles (consisting of vehicle with index 0 to B-1), N is the number of samples for each vechile, T is the number of timesteps (each represents 0.1s), and 6 represents the (x, y, vel, yaw, acc, yawvel) in corresponding agent coordinate of each vehicle. data\_batch is a dictionary that can be used as parameter for relevant APIs. The function should return a loss for every sample of every vehicle with the shape (B, N) or return a loss for every sample with the shape (N)."

% We additionally specify the format and expected return value of the loss function.
% \begin{lstlisting}
% "The generated loss class should have a function \
% forward(x, data_batch, agt_mask). x is a tensor representing the current trajectory with shape (B, N, T, 6) where B is the number of vehicles (consisting of vehicle with index 0 to B-1), N is the number of samples for each vechile, T is the number of timesteps (each represents 0.1s), and 6 represents the (x, y, vel, yaw, acc, yawvel) in corresponding agent coordinate of each vehicle. data_batch is a dictionary that can be used as parameter for relevant APIs. The function should return a loss for every sample of every vehicle with the shape (B, N) or return a loss for every sample with the shape (N).
% \end{lstlisting}

\subsection{\new{Success Examples with Complete Query and Response}}
In this section, we provide five generated success example programs and their queries. For all the examples, in addition to the user query, GPT4 is provided the APIs for the desired loss function and the helper functions as well as two paired examples: acceleration limit and stay on left. The examples help to elucidate two key aspects:
\begin{itemize}
    \item the standard format of a loss class to generate.
    \item how to manipulate different trajectory dimensions using various helper functions (e.g., transform\_coord\_agents\_to\_world, transform\_coord\_world\_to\_agent\_i, etc.) and construct a differentiable loss on the trajectory.
\end{itemize} 

We first show the complete query (including system messages and user query sent to GPT4 API) and response for \textit{GPT Collision} in \Cref{subsubsec:gpt-collision-example}. Next, in \Cref{subsubsec:gpt-keepdistance-example}-\Cref{subsubsec:gpt-lane-following-example}, for four other success exmples, we only show the user query and response as they use the same system messages as \textit{GPT Collision} in \Cref{subsubsec:gpt-collision-example}).

\subsubsection{\new{Success Example: GPT Collision}}
\label{subsubsec:gpt-collision-example}
The OpenAI GPT4 API allows one to specify multiple system messages (which includes the API of the loss function to generation, the APIs of the helper functions, and two paired prompted examples), and one user message (which is essentially the user query) (see \cite{gptapidoc} for more details). Thus, we list each system message and the user message in the following. The qualitative example of using this returned loss function has been shown in \Cref{fig:gpt_example-ctg++-collision}.

System Message 1 - Loss Function API and Helper Functions API:
\begin{python}
    "The generated loss class should have a function \
    forward(x, data_batch, agt_mask). x is a tensor representing the current trajectory with shape (B, N, T, 6) where B is the number of vehicles (consisting of vehicle 0 to B-1), N is the number of samples for each vechile, T is the number of timesteps (each represents 0.1s), and 6 represents the (x, y, vel, yaw, acc, yawvel) in corresponding agent coordinate of each vehicle. data_batch is a dictionary that can be used as parameter for relevant APIs. The function should return a loss for every sample of every vehicle with the shape (B, N) or return a loss for every sample with the shape (N). \
    You can use PyTorch and the following APIs if needed amd you should not use other unseen functions: \
    1. transform_coord_agents_to_world(pos_pred, yaw_pred, data_batch). pos_pred is the predicted position trajectory in agent coordinate with shape (B, N, T, 2) and 2 correspond to (x, y). yaw_pred is the predicted yaw trajectory in agent coordinate with shape (B, N, T, 1). The function transform the predicted position and yaw from their agent coordinates into the world coordinate. The function returns position and yaw in the world coordinate with the shape (B, N, T, 2) and (B, N, T, 1). \
    2. transform_coord_world_to_agent_i(pos_pred_world, yaw_pred_world, data_batch, ind_k). pos_pred is the predicted position trajectory in world coordinate with shape (B, N, T, 2) and 2 represents (x, y). yaw_pred is the predicted yaw trajectory in world coordinate with shape (B, N, T, 1). data_batch is the dictionary mentioned before. ind_k is the index whose agent coordinate will be converted to. The function transform the predicted position and yaw from world coordinate to the agent i coordinate. The function returns position and yaw in the agent i coordinate with the shape (B, N, T, 2) and (B, N, T, 1). \
    3. select_agent_ind(x, i). x has shape (B, N, T, k) where k can be any positive integer and i is a non-negative integer representing the selected index. This function returns the slice of x with index i with shape (N, T, k). \
    4. get_current_lane_projection(pos_pred, yaw_pred, data_batch). pos_pred and yaw_pred have shape (B, N, T, 2) and (B, N, T, 1). They are all in agent coordinate. data_batch is a dictionary mentioned earlier. This function returns the projection of each vehicle predicted trajectory on its current lane in agent coordinate with shape (B, N, T, 3) where 3 represents (x, y, yaw). \
    5. get_left_lane_projection(pos_pred, yaw_pred, data_batch). It is similar to get_current_lane except it returns the left lane waypoints. If there is no left lane, the original trajectory will be returned. \
    6. get_right_lane_projection(pos_pred, yaw_pred, data_batch). It is similar to get_current_lane except it returns the right lane waypoints. If there is no left lane, the original trajectory will be returned."
\end{python}

System Message 2 - Paired Example 1 Query:
\begin{python}
"Generate a loss class such that vehicle 1 should always drive with acceleration below acc_limit."
\end{python}

System Message 2 - Paired Example 1 Response:
\begin{python}
class AccLimitLoss(GuidanceLoss):
    '''
    Keep accelerations below a certain limit.
    '''
    def __init__(self, acc_limit):
        '''
        - acc_limit : acceleration limit.
        '''
        super().__init__()
        self.acc_limit = acc_limit

    def forward(self, x, data_batch, agt_mask=None):
        '''
        - x : the current trajectory (B, N, T, 6) where N is the number of samples and 6 is (x, y, vel, yaw, acc, yawvel)

        - loss: (B, N)
        '''
        if agt_mask is not None:
            x = x[agt_mask]
        acc = x[..., [4]]
        acc_dev = torch.abs(acc) - self.acc_limit
        acc_loss = torch.clip(acc_dev, min=0)
        loss = torch.mean(acc_loss, dim=[-2, -1])

        return loss
\end{python}

System Message 3 - Paired Example 2 Query:
\begin{python}
"Generate a loss class such that vehicle 20 should always stay on the left side of vehicle 13."
\end{python}
System Message 3 - Paired Example 2 Response:
\begin{python}
class StayOnLeftLoss(GuidanceLoss):
    '''
    Vehicle with index target_ind should always keep on the left side of vehicle with index ref_ind.
    '''
    def __init__(self, target_ind=20, ref_ind=13, decay_rate=0.9):
        super().__init__()
        self.target_ind = target_ind
        self.ref_ind = ref_ind
        self.decay_rate = decay_rate

    def forward(self, x, data_batch, agt_mask=None):
        B, N, T, _ = x.shape
        if agt_mask is not None:
            x = x[agt_mask]
        # Select positions
        # (B,N,T,6) -> (B,N,T,2)
        pos_pred = x[..., :2]
        # Select yaws
        # (B,N,T,6) -> (B,N,T,1)
        yaw_pred = x[..., 3:4]
        # convert prediction from the respective agent coordinates to the world coordinate
        # (B,N,T,2), (B,N,T,1), dict -> (B,N,T,2), (B,N,T,1)
        pos_pred_world, yaw_pred_world = 
        transform_coord_agents_to_world(pos_pred, yaw_pred, data_batch)
        # convert prediction from the world coordinate to the agent self.ref_ind coordinate
        # (B,N,T,2), (B,N,T,1), dict, int -> (B,N,T,2), (B,N,T,1)
        pos_pred_in_ref_ind, _ = transform_coord_world_to_agent_i(pos_pred_world, yaw_pred_world, data_batch, self.ref_ind)

        # select the relevant agents with index self.target_ind in the agent self.ref_ind coordinate
        # (B, N, T, 2), int -> (N, T, 2)
        pos_pred_i_in_ref_ind = select_agent_ind(pos_pred_in_ref_ind, self.target_ind)
        # select the relevant agents with index self.target_ind in the agent self.ref_ind coordinate
        # (B, N, T, 2), int -> (N, T, 2)
        pos_pred_j_in_ref_ind = select_agent_ind(pos_pred_in_ref_ind, self.ref_ind)
        
        # Since we only care about the y-axis, we only need to compute the y-axis (rather than both x and y axis) deviation.
        # (N, T, 2), (N, T, 2) -> (N, T)
        pos_dev = pos_pred_j_in_ref_ind[...,1] - pos_pred_i_in_ref_ind[...,1] 
        
        # Clip the position deviation to 0 so that we only penalize the positive deviation
        # (N, T) -> (N, T)
        pos_loss = torch.clip(pos_dev, min=0)

        # penalize early steps more than later steps
        exp_weights = torch.tensor([self.decay_rate ** t for t in range(T)], device=x.device)
        exp_weights /= exp_weights.sum()
        pos_loss = pos_loss * exp_weights[None, :]
        # Take the mean over time
        # (N, T) -> (N)
        pos_loss = pos_loss.mean(-1)

        return pos_loss
\end{python}

User Message - User Query:
\begin{python}
"Generate a loss class such that vehicle 1 should collide with vehicle 2."
\end{python}

Received Message (from GPT4):
\begin{python}
class CollisionLoss(GuidanceLoss):
    '''
    Vehicle with index target_ind should collide with vehicle with index ref_ind.
    '''
    def __init__(self, target_ind=1, ref_ind=2, collision_radius=1.0):
        super().__init__()
        self.target_ind = target_ind
        self.ref_ind = ref_ind
        self.collision_radius = collision_radius

    def forward(self, x, data_batch, agt_mask=None):
        B, N, T, _ = x.shape
        if agt_mask is not None:
            x = x[agt_mask]
        # Select positions
        # (B,N,T,6) -> (B,N,T,2)
        pos_pred = x[..., :2]
        # Select yaws
        # (B,N,T,6) -> (B,N,T,1)
        yaw_pred = x[..., 3:4]
        # convert prediction from the respective agent coordinates to the world coordinate
        # (B,N,T,2), (B,N,T,1), dict -> (B,N,T,2), (B,N,T,1)
        pos_pred_world, yaw_pred_world = 
        transform_coord_agents_to_world(pos_pred, yaw_pred, data_batch)

        # select the relevant agents with index self.target_ind in the world coordinate
        # (B, N, T, 2), int -> (N, T, 2)
        pos_pred_i_world = select_agent_ind(pos_pred_world, self.target_ind)
        # select the relevant agents with index self.ref_ind in the world coordinate
        # (B, N, T, 2), int -> (N, T, 2)
        pos_pred_j_world = select_agent_ind(pos_pred_world, self.ref_ind)

        # Compute the distance between the two vehicles
        # (N, T, 2), (N, T, 2) -> (N, T)
        dist = torch.norm(pos_pred_i_world - pos_pred_j_world, dim=-1)

        # Compute the collision loss by penalizing the distance greater than the collision radius
        # (N, T) -> (N, T)
        collision_loss = torch.clip(self.collision_radius - dist, min=0)

        # Take the mean over time
        # (N, T) -> (N)
        loss = collision_loss.mean(-1)

        return loss
\end{python}

The resulted loss function from GPT4 showcases its capabilities:
\begin{itemize}
\item It adeptly follows the template provided in the examples, producing a syntactically correct program.
\item It comprehends the "collision" concept from the user query and knows how to integrate this concept with the input trajectory, as the examples we used for prompting (acceleration limit and stay on left) are not related to collisions at all. This understanding results in a loss that encourages trajectory collision between two vehicles, achieved through the helper functions. 
\end{itemize}

\subsubsection{\new{Success Example: GPT KeepDistance}}
\label{subsubsec:gpt-keepdistance-example}
In this success example, the returned program indicates that GPT4 understands the meaning of "keep within" and generates a function that promotes vehicle 1 to stay within certain distance of vehicle 2. 

User Message - User Query:
\begin{python}
"Generate a loss class such that vehicle 1 should always keep within 10-30m from vehicle 2."
\end{python}

Received Message (from GPT4):
\begin{python}
class KeepDistanceLoss(GuidanceLoss):
    '''
    Vehicle with index target_ind should always keep within min_distance and max_distance from vehicle with index ref_ind.
    '''
    def __init__(self, target_ind=1, ref_ind=2, min_distance=10, max_distance=30):
        super().__init__()
        self.target_ind = target_ind
        self.ref_ind = ref_ind
        self.min_distance = min_distance
        self.max_distance = max_distance

    def forward(self, x, data_batch, agt_mask=None):
        B, N, T, _ = x.shape
        if agt_mask is not None:
            x = x[agt_mask]
        # Select positions
        # (B,N,T,6) -> (B,N,T,2)
        pos_pred = x[..., :2]
        # Select yaws
        # (B,N,T,6) -> (B,N,T,1)
        yaw_pred = x[..., 3:4]
        # convert prediction from the respective agent coordinates to the world coordinate
        # (B,N,T,2), (B,N,T,1), dict -> (B,N,T,2), (B,N,T,1)
        pos_pred_world, yaw_pred_world = 
        transform_coord_agents_to_world(pos_pred, yaw_pred, data_batch)
        
        # select the relevant agents with index self.target_ind in the world coordinate
        # (B, N, T, 2), int -> (N, T, 2)
        pos_pred_i_world = select_agent_ind(pos_pred_world, self.target_ind)
        # select the relevant agents with index self.ref_ind in the world coordinate
        # (B, N, T, 2), int -> (N, T, 2)
        pos_pred_j_world = select_agent_ind(pos_pred_world, self.ref_ind)

        # Compute the distance between the two vehicles
        # (N, T, 2), (N, T, 2) -> (N, T)
        distance = torch.norm(pos_pred_i_world - pos_pred_j_world, dim=-1)

        # Compute the deviation from the desired distance range
        # (N, T) -> (N, T)
        distance_dev_min = self.min_distance - distance
        distance_dev_max = distance - self.max_distance

        # Clip the negative values to 0
        # (N, T) -> (N, T)
        distance_loss_min = torch.clip(distance_dev_min, min=0)
        distance_loss_max = torch.clip(distance_dev_max, min=0)

        # Combine the losses
        # (N, T) -> (N, T)
        distance_loss = distance_loss_min + distance_loss_max

        # Take the mean over time
        # (N, T) -> (N)
        loss = torch.mean(distance_loss, dim=-1)

        return loss
\end{python}

\subsubsection{\new{Success Example: GPT SameDirection}}
\label{subsubsec:gpt-same-direction-example}
In this success example, the returned program indicates that GPT4 understands the meaning of "same direction" and generates a function that promotes vehicle 1 and vehicle 2 to move with the same orientation. 

User Message - User Query:
\begin{python}
"Generate a loss class such that vehicle 1 should move along the same direction as vehicle 2."
\end{python}

Received Message (from GPT4):
\begin{python}
class SameDirectionLoss(GuidanceLoss):
    '''
    Vehicle with index target_ind should move along the same direction as vehicle with index ref_ind.
    '''
    def __init__(self, target_ind=1, ref_ind=2):
        super().__init__()
        self.target_ind = target_ind
        self.ref_ind = ref_ind

    def forward(self, x, data_batch, agt_mask=None):
        B, N, T, _ = x.shape
        if agt_mask is not None:
            x = x[agt_mask]
        # Select yaws
        # (B,N,T,6) -> (B,N,T,1)
        yaw_pred = x[..., 3:4]
        # convert prediction from the respective agent coordinates to the world coordinate
        # (B,N,T,2), (B,N,T,1), dict -> (B,N,T,2), (B,N,T,1)
        _, yaw_pred_world = transform_coord_agents_to_world(x[..., :2], yaw_pred, data_batch)
        # select the relevant agents with index self.target_ind
        # (B, N, T, 1) -> (N, T, 1)
        yaw_pred_i = select_agent_ind(yaw_pred_world, self.target_ind)
        # select the relevant agents with index self.ref_ind
        # (B, N, T, 1) -> (N, T, 1)
        yaw_pred_j = select_agent_ind(yaw_pred_world, self.ref_ind)

        # Compute the yaw deviation
        # (N, T, 1), (N, T, 1) -> (N, T)
        yaw_dev = torch.abs(yaw_pred_i - yaw_pred_j).squeeze(-1)

        # Normalize the yaw deviation to [0, pi]
        # (N, T) -> (N, T)
        yaw_dev = torch.fmod(yaw_dev, 2 * math.pi)
        yaw_dev = torch.min(yaw_dev, 2 * math.pi - yaw_dev)

        # Penalize deviations larger than a small threshold (e.g., 0.1 radian)
        threshold = 0.1
        yaw_loss = torch.clip(yaw_dev - threshold, min=0)

        # Take the mean over time
        # (N, T) -> (N)
        loss = torch.mean(yaw_loss, dim=-1)

        return loss
\end{python}

\subsubsection{\new{Success Example: GPT Collision From Behind}}
\label{subsubsec:gpt-collision-from-behind-example}
In this success example, the returned program indicates that GPT4 understands the meaning of "collide" and "from behind" and can create the respective losses (see collision\_loss in line 49 and not\_behind\_loss in line 53), respectively. It finally combines the two losses by taking their product as the final loss (line 57).

User Message - User Query:
\begin{python}
"Generate a loss class such that vehicle 1 should collide with vehicle 2 from behind."
\end{python}

Received Message (from GPT4):
\begin{python}
class CollideFromBehindLoss(GuidanceLoss):
    '''
    Vehicle with index target_ind should collide with vehicle with index ref_ind from behind.
    '''
    def __init__(self, target_ind=1, ref_ind=2, collision_threshold=1.0):
        super().__init__()
        self.target_ind = target_ind
        self.ref_ind = ref_ind
        self.collision_threshold = collision_threshold

    def forward(self, x, data_batch, agt_mask=None):
        B, N, T, _ = x.shape
        if agt_mask is not None:
            x = x[agt_mask]
        # Select positions
        # (B,N,T,6) -> (B,N,T,2)
        pos_pred = x[..., :2]
        # Select yaws
        # (B,N,T,6) -> (B,N,T,1)
        yaw_pred = x[..., 3:4]
        # convert prediction from the respective agent coordinates to the world coordinate
        # (B,N,T,2), (B,N,T,1), dict -> (B,N,T,2), (B,N,T,1)
        pos_pred_world, yaw_pred_world = 
        transform_coord_agents_to_world(pos_pred, yaw_pred, data_batch)
        # convert prediction from the world coordinate to the agent self.ref_ind coordinate
        # (B,N,T,2), (B,N,T,1), dict, int -> (B,N,T,2), (B,N,T,1)
        pos_pred_in_ref_ind, _ = transform_coord_world_to_agent_i(pos_pred_world, yaw_pred_world, data_batch, self.ref_ind)

        # select the relevant agents with index self.target_ind in the agent self.ref_ind coordinate
        # (B, N, T, 2), int -> (N, T, 2)
        pos_pred_i_in_ref_ind = select_agent_ind(pos_pred_in_ref_ind, self.target_ind)
        # select the relevant agents with index self.target_ind in the agent self.ref_ind coordinate
        # (B, N, T, 2), int -> (N, T, 2)
        pos_pred_j_in_ref_ind = select_agent_ind(pos_pred_in_ref_ind, self.ref_ind)

        # Compute the distance between the two vehicles along the x-axis (longitudinal)
        # (N, T, 2), (N, T, 2) -> (N, T)
        dist_x = pos_pred_j_in_ref_ind[..., 0] - pos_pred_i_in_ref_ind[..., 0]
        # Compute the distance between the two vehicles along the y-axis (lateral)
        # (N, T, 2), (N, T, 2) -> (N, T)
        dist_y = pos_pred_j_in_ref_ind[..., 1] - pos_pred_i_in_ref_ind[..., 1]

        # Compute the squared distance between the two vehicles
        # (N, T), (N, T) -> (N, T)
        dist_squared = dist_x ** 2 + dist_y ** 2

        # Compute the collision loss by checking if the squared distance is less than the collision threshold
        # (N, T) -> (N, T)
        collision_loss = (dist_squared < self.collision_threshold).float()

        # Penalize the cases where the target vehicle is not behind the reference vehicle
        # (N, T) -> (N, T)
        not_behind_loss = torch.clip(-dist_x, min=0)

        # Combine the collision loss and not_behind_loss
        # (N, T), (N, T) -> (N, T)
        total_loss = collision_loss * not_behind_loss

        # Take the mean over time
        # (N, T) -> (N)
        loss = torch.mean(total_loss, dim=-1)

        return loss
\end{python}

\subsubsection{\new{Success Example: Lane Following}}
\label{subsubsec:gpt-lane-following-example}
In this success example, the returned program indicates that GPT4 can extropolate the number of vehicles of a scenario (although in both provided examples only two vehicles are involved, GPT4 can generalize the loss function involving three vehicles). Besides, GPT4 knows how to use functions described in helper function APIs but not even used in the provided paired examples. In particular, it knows to use get\_current\_lane\_projection to get the lane information in order to promote vehicles lane following behavior when it was asked to let the vehicles to "follow their current lanes".

User Message - User Query:
\begin{python}
"Generate a loss class such that vehicle 1, vehicle 2, and vehicle 3 all follow their current lanes."
\end{python}

Received Message (from GPT4):
\begin{python}
class LaneFollowingLoss(GuidanceLoss):
    '''
    Vehicles with indices target_inds should follow their current lanes.
    '''
    def __init__(self, target_inds=[1, 2, 3]):
        super().__init__()
        self.target_inds = target_inds

    def forward(self, x, data_batch, agt_mask=None):
        B, N, T, _ = x.shape
        if agt_mask is not None:
            x = x[agt_mask]
        # Select positions
        # (B,N,T,6) -> (B,N,T,2)
        pos_pred = x[..., :2]
        # Select yaws
        # (B,N,T,6) -> (B,N,T,1)
        yaw_pred = x[..., 3:4]

        # Get the current lane projection in agent coordinate
        # (B,N,T,2), (B,N,T,1), dict -> (B,N,T,3)
        lane_proj = get_current_lane_projection(pos_pred, yaw_pred, data_batch)

        # Compute the deviation from the lane projection
        # (B,N,T,2), (B,N,T,3) -> (B,N,T,2)
        pos_dev = pos_pred - lane_proj[..., :2]
        yaw_dev = yaw_pred - lane_proj[..., 2:3]

        # Compute the squared loss for position and yaw deviation
        # (B,N,T,2) -> (B,N,T)
        pos_loss = torch.sum(pos_dev ** 2, dim=-1)
        # (B,N,T,1) -> (B,N,T)
        yaw_loss = torch.squeeze(yaw_dev ** 2, dim=-1)

        # Combine position and yaw loss
        # (B,N,T), (B,N,T) -> (B,N,T)
        total_loss = pos_loss + yaw_loss

        # Select the loss for the target vehicles
        # (B,N,T), list -> (len(target_inds), N, T)
        target_losses = [select_agent_ind(total_loss, ind) for ind in self.target_inds]

        # Stack the losses for target vehicles
        # list -> (len(target_inds), N, T)
        target_losses = torch.stack(target_losses, dim=0)

        # Take the mean over time
        # (len(target_inds), N, T) -> (len(target_inds), N)
        target_losses = torch.mean(target_losses, dim=-1)

        # Take the mean over target vehicles
        # (len(target_inds), N) -> (N)
        loss = torch.mean(target_losses, dim=0)

        return loss
\end{python}

\subsection{Failure Cases}
\label{sec:gpt-failure}
The main limitation of the current language interface is on the complex interactions between the vehicles and the map. As we don't explicitly pass the map information into the language interface, it cannot handle commands involving heavy interaction with the map. For example, "vehicle A and B move to the rightmost lane one by one and then both turn right at the next intersection". However, we believe if one provides more helper functions (especially those interacting with the map) and more relevant examples to the language interface, LLM can handle such more complicated commands. \new{Additionally, we observe GPT4 can also fail to understand some concepts properly (e.g., "cut in"). As the current work aims to provide early promising results on the feasibility of text-to-traffic, we leave a deeper exploration for future work.}

\subsubsection{\new{Failure Example : Move to Rightmost and Right Turn}}
\label{subsubsec:gpt-failure-example}
We provide a failure example below. The system messages remain the same as in \Cref{{subsubsec:gpt-collision-example}}. The generated loss function fails because it fails to consider "turn right at the next intersection". This can be potentially fixed by providing a couple of more helper functions (e.g., a function to get the next intersection and corresponding lane's information) and an additional example that interacts with intersection. In other words, our framework can potentially support such queries via extensions on helper functions and additional examples. As in the current work, we mainly want to show a framework of using LLM on language-guided realistic traffic generation, we leave a comprehensive study covering all kinds of complex queries for traffic generation as future work.

User Message - User Query:
\begin{python}
"vehicle 1 and 2 move to the rightmost lane one by one and then both turn right at the next intersection."
\end{python}

Received Message (from GPT4):
\begin{python}
class RightTurnLoss(GuidanceLoss):
    '''
    Vehicle 1 and 2 should move to the rightmost lane one by one and then both turn right at the next intersection.
    '''
    def __init__(self, target_inds=[1, 2], decay_rate=0.9):
        super().__init__()
        self.target_inds = target_inds
        self.decay_rate = decay_rate

    def forward(self, x, data_batch, agt_mask=None):
        B, N, T, _ = x.shape
        if agt_mask is not None:
            x = x[agt_mask]
        # Select positions
        # (B,N,T,6) -> (B,N,T,2)
        pos_pred = x[..., :2]
        # Select yaws
        # (B,N,T,6) -> (B,N,T,1)
        yaw_pred = x[..., 3:4]
        
        # Get the right lane projection
        # (B, N, T, 2), (B, N, T, 1), dict -> (B, N, T, 3)
        right_lane_proj = get_right_lane_projection(pos_pred, yaw_pred, data_batch)

        # Compute the deviation between the predicted position and the right lane projection
        # (B, N, T, 2), (B, N, T, 3) -> (B, N, T, 2)
        pos_dev = pos_pred - right_lane_proj[..., :2]

        # Compute the squared distance
        # (B, N, T, 2) -> (B, N, T)
        pos_dev_sq = torch.sum(pos_dev ** 2, dim=-1)

        # Penalize early steps more than later steps
        exp_weights = torch.tensor([self.decay_rate ** t for t in range(T)], device=x.device)
        exp_weights /= exp_weights.sum()
        pos_dev_sq = pos_dev_sq * exp_weights[None, None, :]

        # Take the mean over time
        # (B, N, T) -> (B, N)
        pos_loss = pos_dev_sq.mean(-1)

        # Select the loss for target vehicles
        # (B, N), list -> (len(target_inds), N)
        target_loss = torch.stack([select_agent_ind(pos_loss, i) for i in self.target_inds])

        # Take the mean over target vehicles
        # (len(target_inds), N) -> (N)
        loss = target_loss.mean(0)

        return loss

\end{python}

\subsubsection{\new{Failure Example: Cut In if Behind and On The Left}}
In this failure example, the returned program indicates that GPT4 fails to understand "cut in" as it only promotes vehicle 1 to be ahead of vehicle 2 when it is behind and on the left. However, cut-in requires vehicle 1 to also move horizontally (i.e., along the y axis).

User Message - User Query:
\begin{python}
"Generate a loss class such that vehicle 1 should cut in ahead of vehicle 2 if it is behind vehicle 2 and on its left lane."
\end{python}

Received Message (from GPT4):
\begin{python}
class CutInLoss(GuidanceLoss):
    '''
    Vehicle with index target_ind should cut in ahead of vehicle with index ref_ind if it is behind vehicle ref_ind and on its left lane.
    '''
    def __init__(self, target_ind=1, ref_ind=2, decay_rate=0.9):
        super().__init__()
        self.target_ind = target_ind
        self.ref_ind = ref_ind
        self.decay_rate = decay_rate

    def forward(self, x, data_batch, agt_mask=None):
        B, N, T, _ = x.shape
        if agt_mask is not None:
            x = x[agt_mask]
        # Select positions
        # (B,N,T,6) -> (B,N,T,2)
        pos_pred = x[..., :2]
        # Select yaws
        # (B,N,T,6) -> (B,N,T,1)
        yaw_pred = x[..., 3:4]
        # convert prediction from the respective agent coordinates to the world coordinate
        # (B,N,T,2), (B,N,T,1), dict -> (B,N,T,2), (B,N,T,1)
        pos_pred_world, yaw_pred_world = 
        transform_coord_agents_to_world(pos_pred, yaw_pred, data_batch)
        # convert prediction from the world coordinate to the agent self.ref_ind coordinate
        # (B,N,T,2), (B,N,T,1), dict, int -> (B,N,T,2), (B,N,T,1)
        pos_pred_in_ref_ind, _ = transform_coord_world_to_agent_i(pos_pred_world, yaw_pred_world, data_batch, self.ref_ind)

        # select the relevant agents with index self.target_ind in the agent self.ref_ind coordinate
        # (B, N, T, 2), int -> (N, T, 2)
        pos_pred_i_in_ref_ind = select_agent_ind(pos_pred_in_ref_ind, self.target_ind)
        # select the relevant agents with index self.target_ind in the agent self.ref_ind coordinate
        # (B, N, T, 2), int -> (N, T, 2)
        pos_pred_j_in_ref_ind = select_agent_ind(pos_pred_in_ref_ind, self.ref_ind)

        # Compute the x-axis and y-axis deviation
        # (N, T, 2), (N, T, 2) -> (N, T)
        pos_dev_x = pos_pred_j_in_ref_ind[..., 0] - pos_pred_i_in_ref_ind[..., 0]
        pos_dev_y = pos_pred_j_in_ref_ind[..., 1] - pos_pred_i_in_ref_ind[..., 1]

        # Check if the target vehicle is behind and on the left lane of the reference vehicle
        behind_and_left = (pos_dev_x > 0) & (pos_dev_y > 0)

        # Penalize the target vehicle for not cutting in ahead of the reference vehicle
        cut_in_loss = torch.where(behind_and_left, pos_dev_x, torch.zeros_like(pos_dev_x))

        # Penalize early steps more than later steps
        exp_weights = torch.tensor([self.decay_rate ** t for t in range(T)], device=x.device)
        exp_weights /= exp_weights.sum()
        cut_in_loss = cut_in_loss * exp_weights[None, :]
        # Take the mean over time
        # (N, T) -> (N)
        cut_in_loss = cut_in_loss.mean(-1)

        return cut_in_loss
\end{python}

\subsection{\new{Qualitative comparison between GPT4 and GPT3.5}}
We next show two examples where we send the same message but use GPT3.5 rather than GPT4 as the LLM. We keep the queries to be exactly the same as in \Cref{subsubsec:gpt-collision-example} and \Cref{subsubsec:gpt-keepdistance-example}, respectively. The received messages are shown in the following. Both loss functions are correct but are more similar to the provided examples by containing code on applying a decay on loss over time dimension (e.g., see line 45-48 in Received Message (from GPT3.5)). These lines of code are almost the exactly ones (except the loss name) in a provided example (stay on left) but these code are not always necessary to be included in the loss functions.

\subsubsection{GPT Collision (GPT3.5)}
User Message - User Query:
\begin{python}
"Generate a loss class such that vehicle 1 should collide with vehicle 2."
\end{python}
Received Message (from GPT3.5):
\begin{python}
class CollisionLoss(GuidanceLoss):
    '''
    Vehicle with index target_ind should collide with vehicle with index ref_ind.
    '''
    def __init__(self, target_ind=1, ref_ind=2, decay_rate=0.9, collision_distance=2.0):
        super().__init__()
        self.target_ind = target_ind
        self.ref_ind = ref_ind
        self.decay_rate = decay_rate
        self.collision_distance = collision_distance

    def forward(self, x, data_batch, agt_mask=None):
        B, N, T, _ = x.shape
        if agt_mask is not None:
            x = x[agt_mask]
        # Select positions
        # (B,N,T,6) -> (B,N,T,2)
        pos_pred = x[..., :2]
        # Select yaws
        # (B,N,T,6) -> (B,N,T,1)
        yaw_pred = x[..., 3:4]
        # convert prediction from the respective agent coordinates to the world coordinate
        # (B,N,T,2), (B,N,T,1), dict -> (B,N,T,2), (B,N,T,1)
        pos_pred_world, yaw_pred_world = 
        transform_coord_agents_to_world(pos_pred, yaw_pred, data_batch)
        # convert prediction from the world coordinate to the agent self.ref_ind coordinate
        # (B,N,T,2), (B,N,T,1), dict, int -> (B,N,T,2), (B,N,T,1)
        pos_pred_in_ref_ind, _ = transform_coord_world_to_agent_i(pos_pred_world, yaw_pred_world, data_batch, self.ref_ind)

        # select the relevant agents with index self.target_ind in the agent self.ref_ind coordinate
        # (B, N, T, 2), int -> (N, T, 2)
        pos_pred_i_in_ref_ind = select_agent_ind(pos_pred_in_ref_ind, self.target_ind)
        # select the relevant agents with index self.target_ind in the agent self.ref_ind coordinate
        # (B, N, T, 2), int -> (N, T, 2)
        pos_pred_j_in_ref_ind = select_agent_ind(pos_pred_in_ref_ind, self.ref_ind)

        # Compute the distance between the two agents
        # (N, T, 2), (N, T, 2) -> (N, T)
        dist = torch.norm(pos_pred_i_in_ref_ind - pos_pred_j_in_ref_ind, dim=-1)

        # Clip the distance to the collision distance
        # (N, T) -> (N, T)
        dist_loss = torch.clip(self.collision_distance - dist, min=0)

        # penalize early steps more than later steps
        exp_weights = torch.tensor([self.decay_rate ** t for t in range(T)], device=x.device)
        exp_weights /= exp_weights.sum()
        dist_loss = dist_loss * exp_weights[None, :]
        # Take the mean over time
        # (N, T) -> (N)
        dist_loss = dist_loss.mean(-1)

        return dist_loss
\end{python}

\subsubsection{GPT Keep Distance (GPT3.5)}
User Message - User Query:
\begin{python}
"Generate a loss class such that vehicle 1 should always keep within 10-30m from vehicle 2."
\end{python}
Received Message (from GPT3.5):
\begin{python}
class KeepDistanceLoss(GuidanceLoss):
    '''
    Vehicle with index target_ind should always keep within min_dist and max_dist from vehicle with index ref_ind.
    '''
    def __init__(self, target_ind=1, ref_ind=2, min_dist=10, max_dist=30, decay_rate=0.9):
        super().__init__()
        self.target_ind = target_ind
        self.ref_ind = ref_ind
        self.min_dist = min_dist
        self.max_dist = max_dist
        self.decay_rate = decay_rate

    def forward(self, x, data_batch, agt_mask=None):
        B, N, T, _ = x.shape
        if agt_mask is not None:
            x = x[agt_mask]
        # Select positions
        # (B,N,T,6) -> (B,N,T,2)
        pos_pred = x[..., :2]
        # Select yaws
        # (B,N,T,6) -> (B,N,T,1)
        yaw_pred = x[..., 3:4]
        # convert prediction from the respective agent coordinates to the world coordinate
        # (B,N,T,2), (B,N,T,1) -> (B,N,T,2), (B,N,T,1)
        pos_pred_world, yaw_pred_world = 
        transform_coord_agents_to_world(pos_pred, yaw_pred, data_batch)
        # convert prediction from the world coordinate to the agent self.ref_ind coordinate
        # (B,N,T,2), (B,N,T,1), dict, int -> (B,N,T,2), (B,N,T,1)
        pos_pred_in_ref_ind, _ = transform_coord_world_to_agent_i(pos_pred_world, yaw_pred_world, data_batch, self.ref_ind)

        # select the relevant agents with index self.target_ind in the agent self.ref_ind coordinate
        # (B, N, T, 2), int -> (N, T, 2)
        pos_pred_i_in_ref_ind = select_agent_ind(pos_pred_in_ref_ind, self.target_ind)
        # select the relevant agents with index self.target_ind in the agent self.ref_ind coordinate
        # (B, N, T, 2), int -> (N, T, 2)
        pos_pred_j_in_ref_ind = select_agent_ind(pos_pred_in_ref_ind, self.ref_ind)
        
        # Compute the distance between the two agents
        # (N, T, 2), (N, T, 2) -> (N, T)
        dist = torch.norm(pos_pred_j_in_ref_ind - pos_pred_i_in_ref_ind, dim=-1)
        
        # Compute the deviation from the desired distance range
        # (N, T) -> (N, T)
        dist_dev = torch.where(dist < self.min_dist, self.min_dist - dist, torch.where(dist > self.max_dist, dist - self.max_dist, torch.zeros_like(dist)))
        
        # penalize early steps more than later steps
        exp_weights = torch.tensor([self.decay_rate ** t for t in range(T)], device=x.device)
        exp_weights /= exp_weights.sum()
        dist_dev = dist_dev * exp_weights[None, :]
        # Take the mean over time
        # (N, T) -> (N)
        dist_loss = dist_dev.mean(-1)

        return dist_loss
\end{python}

\section{\new{Quantitative Evaluation Results with Additional Runs}}
To substantiate the superior performance of our method, we conduct additional experiments, the results of which are presented in \Cref{table:main_result_with_std}. In these experiments, we compare \method against the strongest baseline, CTG, across three distinct runs with varying random seeds. The settings are the same as those for \Cref{table:main_result} and we take the average and standard deviation of the three runs. We highlight the better value only when it is significantly better than the other (i.e., if the values of the two methods differ by at least the sum of their standard deviations). In all eight settings, \method significantly performs better than CTG in terms of failure rate and scene-level realism. \method also tends to perform better than CTG in terms of rule satisfaction (winning 4 and tied on 6). In terms of realism, CTG++ is comparable to CTG (winning 3, losing 2, and tied on 3). Thus, the results suggest that \method significantly performs better than the strongest baseline \method.

\begin{table}[ht]
\centering
\scriptsize
\caption{
    \small{Quantitative results (mean with standard deviation of three runs) of \method and the strongest baselines CTG under GPT-generated rules and STL rules. We highlight the winning method that is significantly better than the other (i.e., if the values of the two methods differ by at least the sum of their standard deviations).}\label{table:main_result_with_std}
}
\resizebox{1.0\textwidth}{!}{
\begin{tabularx}{1.15\textwidth}{l|cccc|cccc}
\toprule
      & \multicolumn{4}{c|}{GPT keep distance}                             & \multicolumn{4}{c}{GPT collision}                                 \\
      \midrule
      & fail             & rule     & real             & rel real           & fail             & rule     & real             & rel real           \\
CTG   & $0.327\pm 0.02$  & $0\pm 0$ & $0.07\pm 0.006$  & $0.343\pm 0.003$ & $0.346\pm 0.018$ & $0\pm 0$ & $\mathbf{0.071\pm 0.004}$ & $0.349\pm 0.002$ \\
CTG++ & $\mathbf{0.171\pm 0.002}$ & $0\pm 0$ & $0.071\pm 0.006$ & $\mathbf{0.334\pm 0.003}$ & $\mathbf{0.264\pm 0.013}$ & $0\pm 0$ & $0.084\pm 0.004$ & $\mathbf{0.336\pm 0.006}$ \\
\end{tabularx}
}

\resizebox{1.0\textwidth}{!}{
\begin{tabularx}{1.25\textwidth}{l|cccc|cccc}
\toprule
      & \multicolumn{4}{c|}{no collision}                                          & \multicolumn{4}{c}{speed limit}                                       \\
      \midrule
      & fail             & rule             & real             & rel real         & fail             & rule         & real             & rel real         \\
CTG   & $0.137\pm 0.01$  & $0.048\pm 0.003$ & $0.048\pm 0.005$ & $0.346\pm 0.002$ & $0.129\pm 0.002$ & $0.029\pm 0$ & $0.077\pm 0.002$ & $0.353\pm 0.003$ \\
CTG++ & $\mathbf{0.085\pm 0.002}$ & $0.045\pm 0.001$ & $0.047\pm 0.007$ & $\mathbf{0.326\pm 0.006}$ & $\mathbf{0.087\pm 0.004}$ & $\mathbf{0.028\pm 0}$ & $\mathbf{0.042\pm 0.003}$ & $\mathbf{0.34\pm 0.004}$  \\
\end{tabularx}
}

\resizebox{1.0\textwidth}{!}{
\begin{tabularx}{1.25\textwidth}{l|cccc|cccc}
\toprule
      & \multicolumn{4}{c|}{target speed}                                          & \multicolumn{4}{c}{no offroad}                                        \\
      \midrule
      & fail             & rule             & real             & rel real         & fail             & rule         & real             & rel real         \\
CTG   & $0.083\pm 0.007$ & $0.281\pm 0.001$ & $0.108\pm 0.003$ & $0.379\pm 0.002$ & $0.167\pm 0.008$ & $0.003\pm 0$ & $0.041\pm 0.002$ & $0.343\pm 0.003$ \\
CTG++ & $\mathbf{0.062\pm 0.001}$ & $\mathbf{0.272\pm 0.002}$ & $\mathbf{0.083\pm 0.004}$ & $\mathbf{0.371\pm 0.004}$ & $\mathbf{0.104\pm 0.008}$ & $0.003\pm 0$ & $0.044\pm 0.006$ & $\mathbf{0.323\pm 0.005}$ \\
\end{tabularx}
}

\resizebox{1.0\textwidth}{!}{
\begin{tabularx}{1.55\textwidth}{l|ccccc|ccccc}
\toprule
      & \multicolumn{5}{c|}{goal waypoint+target speed}                                               & \multicolumn{5}{c}{stopregion+offroad}                                                      \\
      \midrule
      & fail             & rule1            & rule2            & real             & rel real         & fail            & rule1            & rule2            & real             & rel real         \\
CTG   & $0.135\pm 0.015$ & $2.407\pm 0.016$ & $0.39\pm 0.003$  & $0.052\pm 0.002$ & $0.343\pm 0.002$ & $0.128\pm 0.01$ & $0.003\pm 0.001$ & $0.795\pm 0.017$ & $\mathbf{0.046\pm 0.006}$ & $0.336\pm 0.003$ \\
CTG++ & $\mathbf{0.103\pm 0.002}$ & $\mathbf{2.361\pm 0.021}$ & $0.394\pm 0.003$ & $\mathbf{0.039\pm 0.001}$ & $\mathbf{0.336\pm 0.004}$ & $\mathbf{0.08\pm 0.012}$ & $0.003\pm 0$     & $\mathbf{0.44\pm 0.051}$  & $0.076\pm 0.001$ & $\mathbf{0.323\pm 0.006}$ \\
\bottomrule
\end{tabularx}
}
\end{table}

\end{document}